\documentclass{article}

\usepackage[main,final]{neurips_2026}

\usepackage[utf8]{inputenc}
\usepackage[T1]{fontenc}
\usepackage{hyperref}
\usepackage{url}
\usepackage{amsmath}
\usepackage{amssymb}
\usepackage{booktabs}
\usepackage{siunitx}
\usepackage{amsfonts}
\usepackage{multirow}
\usepackage{nicefrac}
\usepackage{microtype}
\usepackage[table]{xcolor}
\usepackage{graphicx}
\usepackage{pifont}
\usepackage{caption}
\usepackage{xspace}

\graphicspath{{img/}}

\definecolor{ourcolor}{HTML}{3366CC}
\newcommand{\ours}{\textcolor{ourcolor}{HOT}\xspace}

\newcommand{\cmark}{\textcolor{green!60!black}{\ding{51}}}
\newcommand{\xmark}{\textcolor{red!85!black}{\ding{55}}}

\title{\ours: Learning Generalizable Hand-Object Tracking without Human Demonstration}
\author{%
  \textbf{Runyi Yu}$^{1,*}$ \quad
  \textbf{Xiaoyi Lin}$^{2,*}$ \quad
  \textbf{Hok Wai Tsui}$^{1,*}$ \quad
  \textbf{Yinhuai Wang}$^{1,*,\dagger}$ \quad
  \textbf{Zhijie Peng}$^{1}$ \\
  \textbf{Hui Zhang}$^{4}$ \quad
  \textbf{Qihan Zhao}$^{1}$ \quad
  \textbf{Ke Fan}$^{5}$ \quad
  \textbf{Miao Li}$^{2}$ \quad
  \textbf{Jie Song}$^{6}$ \\
  \textbf{Jingbo Wang}$^{3}$ \quad
  \textbf{Qifeng Chen}$^{1,\dagger}$ \quad
  \textbf{Ping Tan}$^{1,\dagger}$ \\
  $^{1}$HKUST \quad
  $^{2}$Wuhan University \quad
  $^{3}$Shanghai AI Laboratory \\
  $^{4}$ETH Zurich \quad
  $^{5}$SJTU \quad
  $^{6}$HKUST (GZ) \\
  \texttt{\{ingrid.yu, yinhuai.wang\}@connect.ust.hk} \\
  {\small $^{*}$Equal contribution \quad
  $^{\dagger}$Corresponding authors} \\
  Project page: \url{https://ingrid789.github.io/hot/}
}

\begin{document}

\maketitle

\begin{abstract}
We present a system for learning generalizable hand-object tracking controllers purely from synthetic data, without requiring any human demonstrations.
Our approach makes two key contributions: (1) HOG, a highly efficient Hand-Object data generator that rapidly synthesizes diverse, large-scale, and uniformly distributed trajectories; and (2) HOT, a Hand-Object Tracker that bridges synthetic-to-physical transfer through reinforcement learning and interaction imitation learning. 
Our method extends to diverse object shapes and hand morphologies. 
\ours also enables cross-simulator transfer.
Through extensive evaluations, we show that our approach enables dexterous hands to track challenging, long-horizon sequences including object re-arrangement and agile in-hand reorientation. 
These results represent a significant step toward scalable foundation controllers for manipulation that can learn entirely from simple, scalable synthetic data, breaking the data bottleneck that has long constrained progress in dexterous manipulation. 

\end{abstract}

\begin{center}
\includegraphics[width=\linewidth]{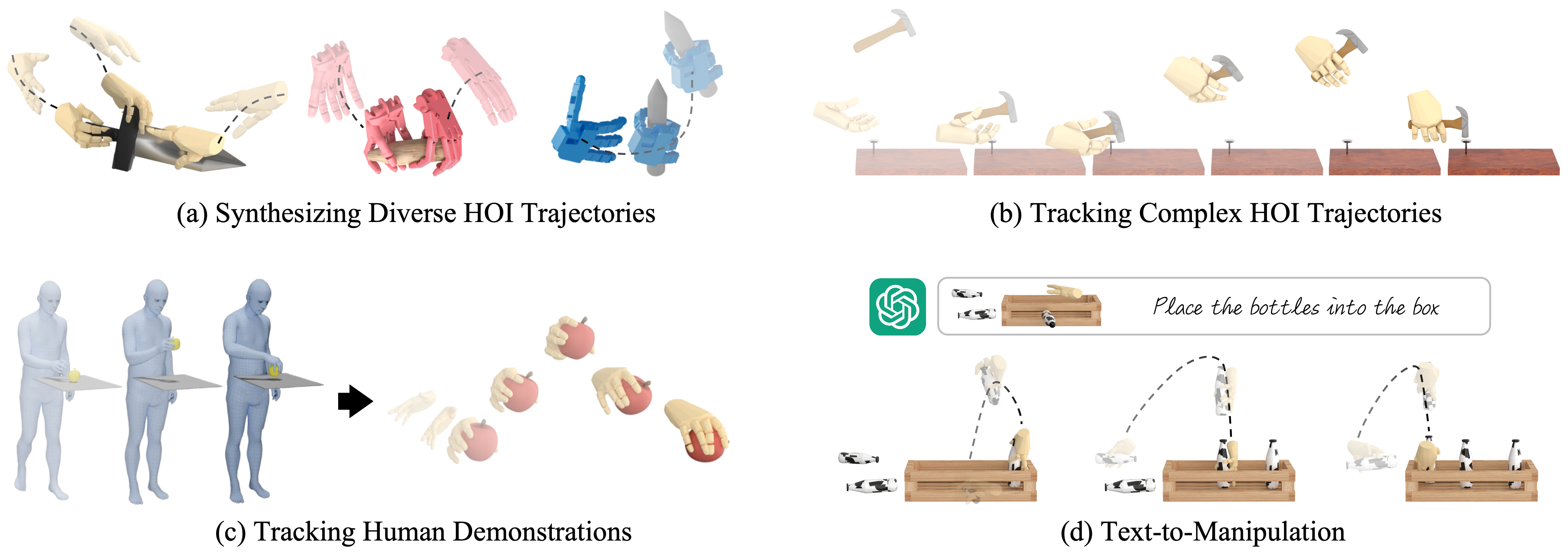}
\par
\refstepcounter{figure}\label{fig: teaser}
\textbf{Figure \thefigure}: Our system synthesizes and tracks hand-object trajectories for a variety of dexterous skills---such as grasping, placing, and in-hand reorientation---as well as their combinations. The approach generalizes across diverse hand and object morphologies and, notably, transfers flexibly to downstream applications including human motion tracking and text-guided task execution. We encourage readers to view our project page for visual results.
\end{center}

\section{Introduction}
\label{sec:intro}

The ability to track and execute complex, dynamic hand-object interactions (HOI) with human-like dexterity is crucial for robots to perform varied dexterous manipulation tasks. 
Traditional teleoperation systems typically track only the robot or hand motion without considering object states, resulting in poor performance for highly dynamic, precision-demanding tasks \citep{cheng2024open,fu2024humanplus}. By incorporating object states into the tracking control loop, hand-object state tracking \citep{wang2023physhoi, wang2024skillmimic, xu2025intermimic, yin2025dexteritygen, liu2025dextrack} significantly enhances the agility and robustness of task execution. 

However, training a generalizable hand-object tracking controller faces the fundamental challenge of acquiring sufficient HOI data.
As summarized in Tab.~\ref{tab:intro_comparison}, existing HOI data sources exhibit different trade-offs in hand generality, scalability, state-space coverage, and reliance on human demonstrations.
Video-based estimation methods such as EgoGrasp~\citep{fu2026egograsp} support scalable data acquisition, but provide limited state-space coverage.
Motion-capture datasets such as GRAB~\citep{GRAB2020} and generative models such as Text2HOI~\citep{cha2024text2hoi} face additional limitations in scalability.
Demonstration augmentation methods such as DexMimicGen~\citep{jiang2024dexmimicgen} improve scalability, while their coverage remains constrained by the source demonstrations.
These representative approaches also remain tied to specific hand representations and depend on human demonstration data.
HOG combines hand-agnostic synthesis, high scalability, and broad state-space coverage without requiring human demonstrations.

In this paper, we address this data challenge by exploring a different route: instead of improving how existing demonstration data are reused, we eliminate the need for human demonstrations in training altogether. 
Our main contribution is therefore systemic and conceptual: we show that a high-performance, generalizable hand-object tracker can be learned from fully synthetic hand-object data, enabling diverse zero-shot applications. Our approach is built on three key insights:
(1) Dexterous manipulation involves continuous transitions between hand-object states, particularly between different grasp configurations. While capturing such continuous interactions is difficult, static states like grasp configurations can be synthesized in large quantities \citep{liu2021synthesizing, zhang2024graspxl, wang2023dexgraspnet}. We capitalize on this asymmetry by generating synthetic manipulation sequences through the transformation, interpolation, and concatenation of these static key frames into plausible pseudo-demonstrations.
%
(2) Although synthetic HOI data may contain physical implausibilities, we find that a policy trained to imitate these HOI data through reinforcement learning can robustly track the corresponding interactions.
(3) A generalizable HOI tracker can be trained effectively on sequences of meta-skills, rather than on long, complete tasks. For instance, instead of modeling ``grasping a hammer and hammering a nail," we synthesize elemental skills like Grasp, Rotate, and Move. This decomposition supports training on short skills; we evaluate zero-shot composition on Grasp--Move--Place sequences, while more complex composites use dedicated training.

Based on these insights, we present HOG (Hand-Object Generator), a scalable manipulation data synthesis framework for generating diverse hand-object trajectories across various object shapes and hand morphologies, and HOT (Hand-Object Tracker), a tracking-based controller trained on these demonstrations, capable of tracking complex, highly dynamic, and long-horizon manipulation trajectories during inference. The key role of HOG is not merely to generate more data, but to produce data that are most conducive to scaling and generalization: supporting diverse objects, diverse hands, diverse skills, and a wide range of grasping and trajectory distributions.
HOG generates synthetic meta-skill trajectories through a structured, two-stage process. First, given a rigid object and a dexterous hand model, it employs force-closure optimization and reinforcement learning \citep{wang2023dexgraspnet, zhang2024graspxl} to generate and refine a diverse set of stable grasp configurations. Subsequently, HOG anchors itself to these grasp poses to synthesize trajectories for fundamental meta-skills—such as Grasping, Moving, Placing, and Rotating—as well as their combinations.
\ours is trained primarily on the meta-skill data synthesized by HOG, using a unified set of HOI imitation rewards. To enhance model capacity and learning efficiency, we introduce a two-stage teacher-student training framework, coupled with a data randomization curriculum that progressively increases the randomness of training samples—such as varying object sizes and initial poses—as the model converges, thereby improving overall robustness.

Our core benchmark compares DexTrack-style specialists, \ours teachers, and one unified seven-skill \ours student on the same HOG data and four objects. We use a separately trained Grasp/Move/Place student to evaluate object generalization after training on 15 objects and zero-shot composition on four-object Grasp--Move--Place trajectories. Beyond downstream tracking performance, we show that HOG occupies a favorable point in the data-design space compared with video-estimated trajectories such as EgoGrasp~\citep{fu2026egograsp}, motion-capture datasets, generative HOI models, and augmentation-based pipelines: it preserves physical structure while achieving substantially better scalability and state-space coverage. Notably, we find that the fast and intentionally noisy generation process of HOG is crucial for endowing \ours with robust interaction dynamics. With dedicated composite-task training, \ours also tracks long-horizon dynamic sequences, such as catching a thrown hammer, reorienting it in midair, and performing a hammering motion. Moreover, \ours shows effectiveness in tracking real human demonstrations \citep{GRAB2020} and data from HOI generative models \citep{cha2024text2hoi}. Notably, the system enables zero-shot transfer to downstream applications. When integrated with LLMs or VLMs \citep{achiam2023gpt, team2023gemini}, it follows language instructions to control both HOG-based data synthesis and \ours’s tracking for manipulation.
%

\begin{table}[t]
  \caption{Comparison of HOI data sources. 
    Unlike existing pipelines that are often constrained by human demonstrations, specific hand morphologies, or narrow state-space coverage, HOG is hand-agnostic and offers high scalability and broad state-space coverage. 
    This makes it an ideal framework for training highly generalizable hand-object trackers.}
  \centering
  \setlength{\tabcolsep}{7pt}
  \renewcommand{\arraystretch}{1.2}
  \resizebox{\linewidth}{!}{
    \begin{tabular}{lccccc}
      \toprule
      \textbf{Data Source \textbackslash{} Property} & \textbf{Method} & \textbf{Hand-Agnostic} & \textbf{Scalability} & \textbf{State-Space Coverage} & \textbf{w/o Human Demo.} \\
      \midrule
      \textbf{Video Estimation}  & EgoGrasp & \xmark & \cmark & \xmark & \xmark \\
      \textbf{Generative Model} & Text2HOI & \xmark & \xmark & \xmark & \xmark \\
      \textbf{Dataset}  & GRAB & \xmark & \xmark & \xmark & \xmark \\
      \textbf{Data Augmentation}  & DexMimicGen & \xmark & \cmark & \xmark & \xmark \\
      \midrule
      \textbf{HOG (Ours)} & - & \cmark & \cmark & \cmark & \cmark \\
      \bottomrule
    \end{tabular}
  }
  \label{tab:intro_comparison}
\end{table}

\vspace{-4pt}\section{Related Work}
\paragraph{Reinforcement Learning for Dexterous Manipulation.}
Recent years have witnessed significant progress in dexterous manipulation, particularly in grasping tasks \citep{chen2023sequential,huang2023dynamic,chen2023bi,yin2023rotating,andrychowicz2020learning,fang2023anygrasp,allshire2022transferring,rajeswaran2018learning,she2022learning, she2024learning}. By leveraging advances in reinforcement learning \citep{schulman2017proximal} and simulation technology \citep{makoviychuk2021isaac,todorov2012mujoco}, current grasp models can handle objects of varying shapes \citep{zhang2024graspxl,xu2023unidexgrasp,wan2023unidexgrasp++}. However, these approaches typically rely on task-specific reward engineering, limiting their generalization to broader manipulation tasks \citep{chen2023sequential}. While similar paradigms have shown impressive results in specific dexterous tasks like object reorientation \citep{andrychowicz2020learning} and articulated object manipulation \citep{zhang2024artigrasp}, they struggle to scale to general manipulation scenarios. Some methods attempt to enhance task diversity through a mixture of experts, but they still cannot circumvent the labor-intensive nature of task-specific reward engineering \citep{yang2020multi}. Recent studies~\citep{wang2023physhoi,wang2024skillmimic, xu2025intermimic, liu2025dextrack, chen2024object} propose a paradigm shift by introducing general imitation rewards to mimic robot-object state trajectories, enabling unified learning across different manipulation tasks. Notably, this paradigm demonstrates impressive robustness to data imperfections, substantially reducing requirements for demonstration quality and quantity. However, acquiring diverse manipulation data remains a significant challenge.

\paragraph{Demonstration Data for Dexterous Manipulation.}
Data collection remains a critical challenge for learning dexterous manipulation skills. Current approaches primarily fall into two categories: human demonstrations and synthetic data generation. 

Human demonstrations are typically collected through teleoperation, device-based motion capture, and video-based estimation. While teleoperation provides accurate action data, it suffers from latency and perception barriers, making agile and precise manipulations challenging \citep{fu2024humanplus,he2024omnih2o,cheng2024open}. Motion capture systems using specialized sensor gloves and camera arrays can capture detailed hand-object motion \citep{banerjee2024introducing, GRAB2020,fan2023arctic}, but incur substantial time and labor costs and remain tied to specific embodiments, objects, and recording setups. Recent advances in video-based pose estimation enable extraction of hand-object motion from vast video datasets \citep{cao2021reconstructing,wen2023bundlesdf,wang2023physhoi}; however, these methods often struggle with occlusion, depth ambiguity, and unstable contact reconstruction, which limits their use as large-scale supervisory signals for precise HOI tracking.

Synthetic data generation has gained increasing attention recently \citep{mandlekar2023mimicgen,jiang2024dexmimicgen,garrett2024skillmimicgen,gao2024efficient,pumacay2024colosseum}, with methods mainly focusing on transforming and recombining existing reference data \citep{mandlekar2023mimicgen,jiang2024dexmimicgen,garrett2024skillmimicgen}. This improves combinatorial diversity, but it does not remove the underlying dependence on human demonstrations, and thus cannot escape the coverage ceiling imposed by the source corpus. HOI generative models are another promising direction, yet current systems still face difficulties in preserving physical validity, stable grasping, and broad state-space coverage for dexterous tracking supervision \citep{cha2024text2hoi}. 

Our work instead focuses on how the training data distribution shapes generalization in HOI tracking. HOG is designed to directly control this distribution by synthesizing physically grounded and scalable trajectories without human demonstrations, overcoming the limited coverage of captured or augmented data and the weak physical reliability of current generative models.


\vspace{-4pt}\section{Method}
Our system consists of two parts: HOG (Hand-Object Generator), a scalable data generation framework, and \ours (Hand-Object Tracker), a tracking-based controller trained on synthetic data, capable of tracking complex, highly dynamic, and long-horizon manipulation trajectories during inference. 
In applications, our system follows a modular pipeline. First, a high-level directive—such as a text command processed by a VLM, a trajectory from an HOI generative model, or a captured human demonstration—is converted into sparse waypoints. These waypoints are then fed to HOG to generate a dense hand-object trajectory, which \ours executes in simulation. This workflow enables zero-shot adaptation to diverse downstream applications.
Fig.~\ref{fig: overview} shows the overall system architecture.

\subsection{Hand-Object Generator}

\label{sec: data gen}

\begin{figure*}[t]
  \centering  \includegraphics[width=0.9\linewidth]{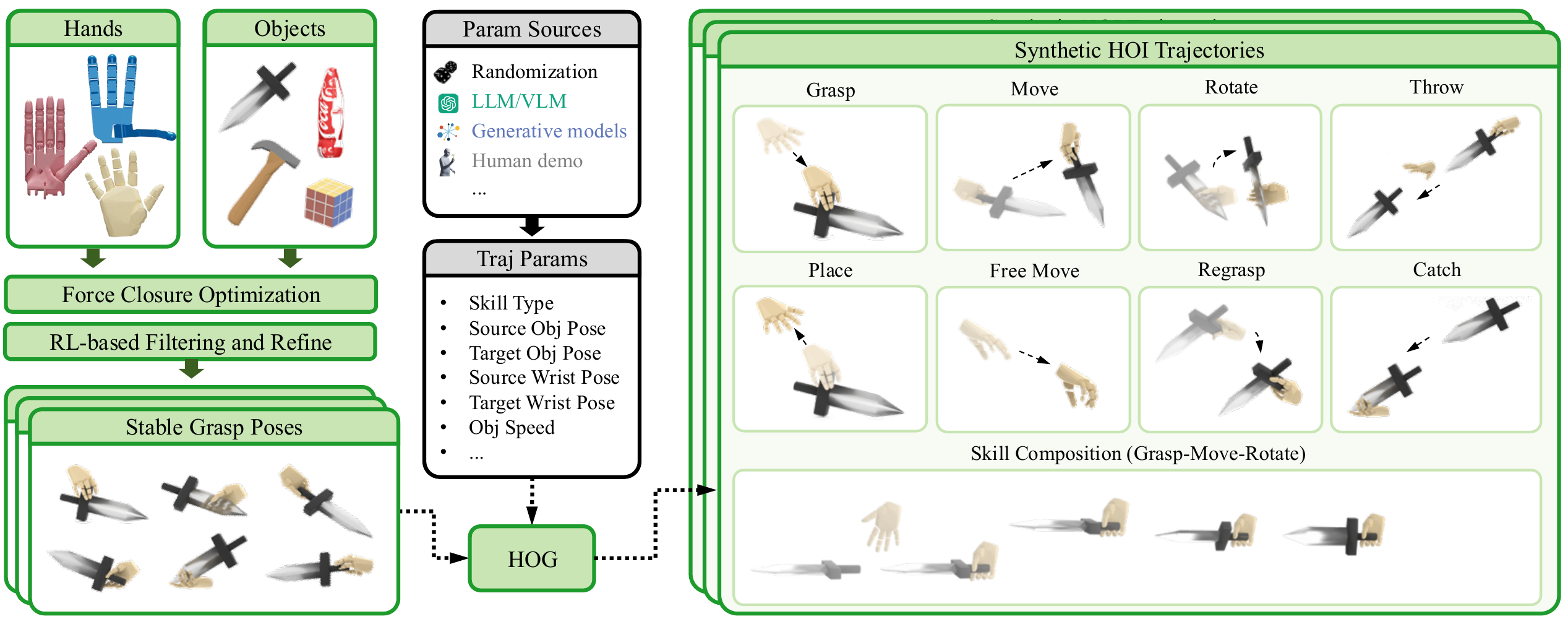}
  \caption{
    HOG synthesizes manipulation trajectories from grasp poses generated by force-closure optimization and refined by RL. It supports 8 meta-skills (7 HOI skills + Free Move) and multiple parameter sources, including randomization, LLM/VLM instructions, and human demonstrations, while naturally generalizing across diverse hands and objects.
  } 
\label{fig: autodexgen}
\vspace{-5pt}
\end{figure*}

The Hand-Object Generator (HOG) consists of two steps: first, synthesizing stable grasp configurations, and second, generating meta-skill demonstrations or their combinations based on these grasp configurations. Fig.~\ref{fig: autodexgen} illustrates our data generation pipeline. 

\subsubsection{Synthesizing Grasp Pose Candidates}
Static grasp configurations can be easily obtained from existing datasets \citep{wang2023dexgraspnet,chao2021dexycb,hasson2019learning,GRAB2020,fan2023arctic}, grasp models \citep{xu2023unidexgrasp,wan2023unidexgrasp++,zhang2024graspxl,miller2004graspit}, or through optimization \citep{wang2023dexgraspnet,liu2021synthesizing}, thereby ensuring substantial scalability potential. 
In this study, we leverage an off-the-shelf grasp generator \citep{wang2023dexgraspnet} driven by differentiable force closure estimation \citep{liu2021synthesizing} to generate diverse grasp configurations. Specifically, given a rigid object and a dexterous hand, we obtain a set of grasp configurations $\hat{\mathcal{G}} = \{\boldsymbol{g}_1, \boldsymbol{g}_2, ..., \boldsymbol{g}_M\}$. The grasp configuration is parameterized by the hand-object state
\begin{equation}
\boldsymbol{g} = \{\mathbf{T}^h, \mathbf{R}^h, \boldsymbol{\theta}, \mathbf{P}^h, \mathbf{T}^o, \mathbf{R}^o, \mathbf{P}^o\}.
\label{eq: grasp configuration}
\end{equation}
For the MANO hand model \citep{romero2022embodied}, $\mathbf{T}^h \in \mathbb{R}^3$ and $\mathbf{R}^h \in SO(3)$ represent the global wrist pose, while $\boldsymbol{\theta} \in \mathbb{R}^{15\times3}$ describes the joint angles of the five fingers (each finger has three joints with three degrees of freedom). $\mathbf{P}^h \in \mathbb{R}^{15\times3}$ denotes the global positions of finger joints. Similarly, $\mathbf{T}^o$, $\mathbf{R}^o$, and $\mathbf{P}^o\in \mathbb{R}^{N\times3}$ represent the object's global pose and the global positions of $N$ keypoints on the object surface.

\subsubsection{Filtering and Refining Grasp Poses}
After obtaining a large set of initial grasp pose candidates, we train a coach policy to filter and refine them into physically plausible and stable configurations for subsequent data synthesis. Specifically, the policy is trained in simulation by initializing the hand and object according to each candidate pose. The coach policy is optimized using a grasp reward that encourages successful grasping and adherence to the original candidate pose. To enhance robustness, we apply random force perturbations to the object during training. Additional implementation details of the coach policy follow the setup described in Sec.~\ref{sec: hot}. Once the policy converges, we collect the successful grasp poses generated during its rollouts as the stable grasp pose set $\mathcal{G}$. These refined poses are significantly higher in quality, physical realism, and stability compared to the original candidates.

\subsubsection{Synthesizing Meta-Skill Trajectories}
\label{sec: traj synth}
A dexterous manipulation trajectory is represented as $\boldsymbol{\tau} = \{\boldsymbol{g}_{t}\}_{t=0}^{T}$, with each $\boldsymbol{g}_{t}$
 parameterized as in Eq.~\ref{eq: grasp configuration}. We synthesize pseudo-demonstrations by interpolating randomly generated hand-object states and grasp configurations. While these trajectories may be physically approximate, they serve as tracking targets in reinforcement imitation learning, enabling the policy to learn generalizable target-action mappings. For efficient synthesis and learning, we decompose manipulation into 7 meta-skills: Grasp, Place, Move, Catch, Throw, Rotate, and Regrasp. Below, we detail the synthesis of Grasp, Move, Place, and Rotate; see the Appendix for the others.

\begin{figure*}[t]
  \centering  \includegraphics[width=1\linewidth]{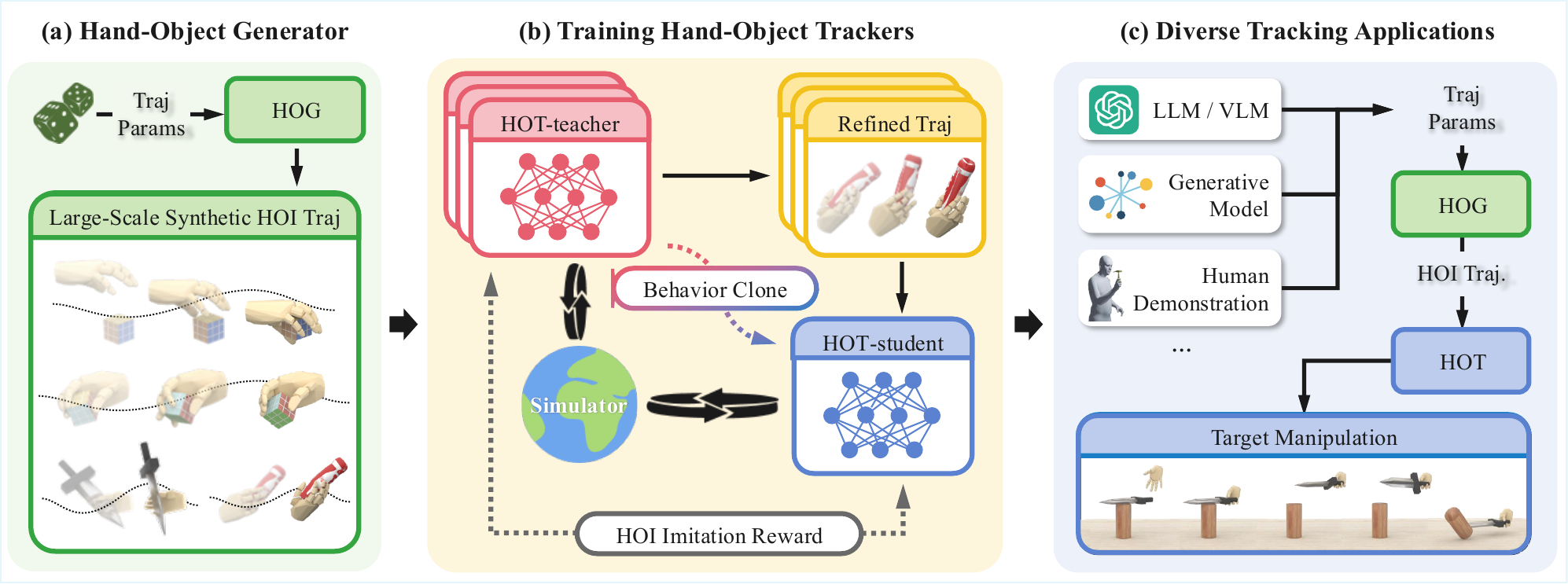}
  \caption{
Our system learns generalizable hand-object tracking from synthetic data. \textbf{(a) HOG} synthesizes manipulation trajectories for meta-skills. \textbf{(b) \ours} is trained through a two-stage teacher-student framework using reinforcement learning with a unified HOI imitation reward, enabling robust tracking of the target HOI trajectories. \textbf{(c) At inference}, the system can accept high-level waypoints from language models, generative models, or human data, which HOG converts into trajectories for \ours to track, enabling diverse applications.
}
\label{fig: overview}
\vspace{-5pt}
\end{figure*}

\vspace{-0.05cm}
\textit{\textbf{Grasp.}} The Grasp skill involves grasping objects placed on the ground. First, we determine the set of stable object poses for a given object through physics simulation by randomly dropping objects and recording their resting poses, denoted as $\mathcal{H} = \{\mathbf{h}_1, \mathbf{h}_2, ..., \mathbf{h}_K\}$, where each $\mathbf{h} = [\mathbf{T}^o, \mathbf{R}^o]$ represents a stable object pose. To generate a grasp trajectory, we sample a grasp configuration $\boldsymbol{g}$ from $\mathcal{G}$ and a stable object pose $\mathbf{h}$ from $\mathcal{H}$, then transform $\boldsymbol{g}$ to align the object pose with $\mathbf{h}$, obtaining $\boldsymbol{g}^{end}$. We reject and resample $\boldsymbol{g}^{end}$ if any hand keypoints penetrate the ground plane; otherwise, we keep it as the grasping frame. To generate the initial state $\boldsymbol{g}^{init}$, we compute the vector $\mathbf{v}$ from the object center to the first joint of the index finger, randomly sample the initial hand position within a cone centered along $\mathbf{v}$, and set all hand joint angles to zero while keeping other parameters identical to $\boldsymbol{g}^{end}$. Finally, we generate the continuous grasp trajectory by interpolating between $\boldsymbol{g}^{init}$ and $\boldsymbol{g}^{end}$.

\vspace{-0.05cm}
\textit{\textbf{Place.}}
Place represents the inverse operation of Grasp: an object is placed stably on a surface and then released by the hand. We efficiently generate Place demonstrations by time-reversing the Grasp trajectories.

\vspace{-0.05cm}
\textit{\textbf{Move.}}
Move involves moving an object through space while maintaining a fixed grasp configuration. To synthesize these demonstrations, we first randomly select a grasp configuration $\boldsymbol{g}$ from $\mathcal{G}$. We then generate random initial and final object poses within the operational space and create an object trajectory through interpolation. The wrist pose is subsequently aligned with the object trajectory at each timestep, maintaining the constant relative grasp configuration throughout the motion. Notably, we also generate Move trajectories starting from the final frame of Grasp demonstrations or ending at the initial frame of Place demonstrations to ensure smooth cross-skill transitions.

\vspace{-0.05cm}
\textit{\textbf{Rotate.}}
Rotate demonstrates in-hand manipulation where the object is rotated while maintaining a fixed wrist pose. We provide details for generating general Rotate data in the appendix.
However, due to its difficulty and lengthy training requirements, we propose a simplified version that balances practical value with complexity. First, we define a rotatable region for each object, such as the handle of a hammer, and an associated rotation axis. If a grasp configuration $\boldsymbol{g}^1$ from $\mathcal{G}$ contacts the object's rotatable region, we designate it as the first keyframe. We then rotate the object by a random angle around the defined axis to create the second keyframe, and continue this process for subsequent keyframes. 
To accommodate large state changes, we replicate each $\boldsymbol{g}_i$ for $N_i$ frames, with $N_i$ scaled by the magnitude of the state difference. This provides a sufficient learning buffer for the policy.

The resulting demonstration is:
\begin{equation}
\begin{aligned}
\{\underset{N_1}{\underbrace{\boldsymbol{g}_{1}, ..., \boldsymbol{g}_{1}}}, \underset{N_2}{\underbrace{\boldsymbol{g}_{2}, ..., \boldsymbol{g}_{2}}}, ..., \underset{N_k}{\underbrace{\boldsymbol{g}_{k}, ..., \boldsymbol{g}_{k}}}\},
\label{eq: sampled traj}
\end{aligned}
\end{equation}

\subsection{Hand-Object Tracker}

\label{sec: hot}
The final goal is to train a Hand-Object Tracker (HOT) that can not only track various meta-skill demonstrations but also track composite long-horizon sequences. We employ reinforcement learning in physical simulation with a unified imitation reward to learn from synthetic demonstrations generated using the procedure in Sec.~\ref{sec: data gen}. We condition the controller on target hand-object states, enabling the controller to accomplish tasks of varying duration and semantic complexity through tracking. 
Further, we improve the tracking performance through a two-stage teacher-student training framework.

\subsubsection{HOT Structure}

We employ a single policy that outputs controls for both the hand's native joints (e.g., MANO's 45 DoFs) and the wrist's 6-DoF pose, which is essential for dynamic tasks. To enhance generalization, we leverage the translational invariance of manipulation: the relative control strategy should be location-agnostic. Consequently, our policy is trained to output residual adjustments for the wrist, ensuring generalization across the workspace.

\ours is formulated as a policy parameterized by an MLP. 
The policy output is parameterized as a Gaussian distribution \begin{equation}
    \boldsymbol{a}_{t} \sim \mathcal{N}(\boldsymbol{\phi}_{\boldsymbol{\pi}}(\boldsymbol{h}_{t}), \boldsymbol{\Sigma}_{\boldsymbol{\pi}}),
    \label{eq: action}
\end{equation}
where $\boldsymbol{\phi}_{\boldsymbol{\pi}}$ is a three-layer MLP that maps input $\boldsymbol{h}_{t}$ to action means. The variances $\boldsymbol{\Sigma}_{\boldsymbol{\pi}}$ are set to non-zero constants during training for exploration and to zero during testing for stability. 
The action $\boldsymbol{a}_{t}$ consists of a residual 6-DoF wrist pose $\boldsymbol{\delta}^{wrist}_{t}$ and target finger joint rotations $\boldsymbol{a}^{finger}_{t}$. For wrist pose control, we employ an incremental control scheme where the target 6-DoF wrist pose is updated as $\boldsymbol{a}^{wrist}_{t} = \boldsymbol{a}^{wrist}_{t-1} + \boldsymbol{\delta}^{wrist}_{t}$. For finger joints, we apply direct position control using $\boldsymbol{a}^{finger}_{t}$ as the target joint rotations. These targets are converted to torque commands through PD controllers.
%

The observation of the \ours policy is
\begin{equation}
    \boldsymbol{h}_{t}=\{\boldsymbol{s}_{t},\hat{\boldsymbol{s}}^{goal}_{t+1}\}
\end{equation}
where $\boldsymbol{s}_{t} = \{\boldsymbol{s}^{hand}_{t},\boldsymbol{s}^{obj}_{t}\}$ is the environmental observation, consisting of the current hand state $\boldsymbol{s}^{hand}_{t}$ and object state $\boldsymbol{s}^{obj}_{t}$ in the simulation environment. The hand state $\boldsymbol{s}^{hand}_{t}$ encompasses the global wrist position $\boldsymbol{s}^{wp}_{t}$ and rotation $\boldsymbol{s}^{wr}_{t}$, finger joint angles $\boldsymbol{s}^{fr}_{t}$ and angular velocities $\boldsymbol{s}^{frv}_{t}$, finger joint positions $\boldsymbol{s}^{fp}_{t}$ and velocities $\boldsymbol{s}^{fpv}_{t}$, and fingertip force sensor readings $\boldsymbol{s}^{fs}_{t}$. The object state $\boldsymbol{s}^{obj}_{t}$ consists of the object position $\boldsymbol{s}^{op}_{t}$, rotation $\boldsymbol{s}^{or}_{t}$, velocity $\boldsymbol{s}^{opv}_{t}$, angular velocity $\boldsymbol{s}^{orv}_{t}$, and the positions of keypoints on the object surface $\boldsymbol{s}^{okp}_{t}$. $\hat{\boldsymbol{s}}^{goal}_{t+1}$ denotes the goal state for the subsequent time step, which encompasses multiple components:
\begin{equation}
    \hat{\boldsymbol{s}}^{goal}_{t+1} = \{\hat{\boldsymbol{s}}^{wr}_{t+1}, \hat{\boldsymbol{s}}^{wp}_{t+1}, \hat{\boldsymbol{s}}^{wpv}_{t+1}, \hat{\boldsymbol{s}}^{fr}_{t+1}, \hat{\boldsymbol{s}}^{op}_{t+1}, \hat{\boldsymbol{s}}^{or}_{t+1}\}
\end{equation}
Here, $\hat{\boldsymbol{s}}^{wpv}_{t+1}$ denotes the reference wrist linear velocity. The remaining symbols follow the notation above, and all goal state values are extracted from the next frame of reference data. Notably, all positional measurements (except for the global wrist position) are transformed into a local coordinate frame attached to the wrist.

\subsubsection{Training the Hand-Object Tracker}
\label{sec: training}

As illustrated in Fig.~\ref{fig: overview}(b), we train the Hand-Object Tracker (HOT) using a two-stage reinforcement learning framework guided by synthetic demonstration trajectories with unified HOI imitation rewards. 
Training starts by sampling a HOG-generated trajectory $\tau = \{\hat{\boldsymbol{s}}_{t}\}_{t=0}^{T}$ (Sec.~\ref{sec: data gen}) and a random starting index $t_0$. The hand-object system is initialized from $\hat{\boldsymbol{s}}_{t_0}$ with perturbations (Appendix~\ref{sec:dr}), and the policy tracks the subsequent reference states.
At each timestep, the policy outputs actions (Eq.~\ref{eq: action}) that are executed in simulation, after which an imitation reward $r_{t} = f(\boldsymbol{s}_{t+1},\hat{\boldsymbol{s}}_{t+1})$ (Sec.~\ref{sec: imitation reward}) is computed by comparing the current state with the reference. Network parameters $\boldsymbol{\phi}_{\boldsymbol{\pi}}$ are optimized using PPO \citep{schulman2017proximal}.


\label{sec: teacher-student}
Although a single policy is effective for individual skills, it faces scalability limitations when learning diverse skills or object categories. The primary challenge arises from physical implausibilities in synthetic data, which require extensive sampling to identify feasible solutions for each instance. As the dataset expands, sampling density per trajectory decreases, adversely affecting convergence.
To overcome this bottleneck, we adopt a two-stage teacher-student distillation framework \citep{he2024omnih2o, luo2023perpetual}. In the first stage, we train specialist teachers for each skill--object pair in the seven-skill benchmark, or object-specific teachers covering Grasp, Move, and Place in the multi-object experiment.
In the second stage, one student learns the skills and objects covered by the corresponding teacher set, using the teacher policies for both online data refinement and action guidance through behavior cloning \citep{juravsky2024superpadl, xu2025intermimic}. We train the seven-skill/four-object student and the three-skill/15-object student separately; each consolidates its own teacher set into one policy.



\subsubsection{Unified Hand-Object Imitation Reward}
\label{sec: imitation reward}

We apply a unified imitation reward to learn manipulation from demonstrations. The reward combines complementary components multiplicatively:
\begin{equation}
\begin{aligned}
r_{t} = r_{t}^{hand}*r_{t}^{wrist}*r_{t}^{object}*r_{t}^{interact}*r_{t}^{contact},
\end{aligned}
\label{eq: imitation reward}
\end{equation}
where $r_{t}^{hand}$ and $r_{t}^{wrist}$ are the hand and wrist tracking rewards that encourage following demonstrated global hand motion trajectories, $r_{t}^{object}$ captures global object motion, $r_{t}^{interact}$ highlights hand-object relative motion, and $r_{t}^{contact}$ encourages correct per-frame contact status.
Given the degradation in our demonstration data, we deliberately exclude velocity terms from the imitation objectives to ensure robust learning. Details about the sub-rewards can be found in the appendix.

\vspace{-4pt}\section{Experiments}

\subsection{Experimental Settings}
\label{metrics}

\paragraph{Settings} Experiments use Isaac Gym~\citep{makoviychuk2021isaac} and one RTX 4090 GPU, with simulation control at 120 Hz and policies at 60 Hz. \ours teachers train for 30,000 epochs and students for 20,000; DexTrack-style uses $3\times20{,}000$ epochs (Appendix~\ref{sec:method-details}). The initial HOG data are shared, but compute is not matched.

\paragraph{Baseline methods} We compare \ours against four baselines: Open-Loop, SkillMimic-style~\citep{wang2024skillmimic}, DexGen-var~\citep{yin2025dexteritygen}, and DexTrack-style~\citep{liu2025dextrack}. (1) Open-Loop is a controller that directly drives a PD controller using joint rotations from reference demonstrations. (2) SkillMimic uses an HOI imitation method that observes only the 6D object pose without shape information. (3) DexGen-var is our reimplementation of DexGen following its original design, inferring object states indirectly from PD controller errors without direct object observations. (4) DexTrack-style implements DexTrack's homotopy optimization within our codebase, refining reference trajectories over three rounds using successful policy rollouts. See details in Appendix~\ref{sec:method-details}.

\paragraph{Metrics} We measure object position error $E_{op}$ (cm), object rotation error $E_{or}$ (degrees), hand mean per-joint position error $E_h$ (cm), and frame-wise success rate (SR). A frame is considered successful when both $E_{op}$ and $E_h$ fall below a skill-specific threshold: 10 cm for Grasp, Move, and Place, and 20 cm for Throw, Catch, Rotate, and Regrasp. Tab.~\ref{tab:main} reports means across four objects (box, bottle, sword, and hammer) and five seeds, with standard deviations across seeds reported only for SR. Each seed uses 1,000 trials per object–skill pair.

\begin{table}[t]
  \centering
  \caption{\textbf{Baseline comparison and reward ablations on meta-skills.} Reward ablation results are reported for the teacher policies. Bold and underlined values indicate the best and second-best distinct values in each column across all listed methods, respectively; ties share the same formatting. See Sec.~\ref{metrics} for metric definitions.}
  \label{tab:main}
  \renewcommand{\arraystretch}{1.1}
  \setlength{\tabcolsep}{3pt}
  \small
  \sisetup{mode=text,detect-weight=true}
  \newcommand{\srnum}[1]{\tablenum[table-format=2.1]{#1}}
  \newcommand{\srstd}[1]{\ensuremath{^{\pm #1}}}
  \newcommand{\errnum}[1]{\tablenum[table-format=2.2]{#1}}
  \resizebox{\linewidth}{!}{%
    \begin{tabular}{l@{\hspace{8pt}}|cccc@{\hspace{8pt}}cccc@{\hspace{8pt}}cccc@{\hspace{8pt}}cccc}
      \toprule
      \multirow{2}{*}{Method} & \multicolumn{4}{c}{\textit{Grasp}} & \multicolumn{4}{c}{\textit{Move}} & \multicolumn{4}{c}{\textit{Place}} & \multicolumn{4}{c}{\textit{Rotate}} \\
      \cmidrule(lr){2-5}\cmidrule(lr){6-9}\cmidrule(lr){10-13}\cmidrule(lr){14-17}
        & SR (\%) $\uparrow$ & $E_{op}\downarrow$ & $E_{or}\downarrow$ & $E_h\downarrow$ & SR (\%) $\uparrow$ & $E_{op}\downarrow$ & $E_{or}\downarrow$ & $E_h\downarrow$ & SR (\%) $\uparrow$ & $E_{op}\downarrow$ & $E_{or}\downarrow$ & $E_h\downarrow$ & SR (\%) $\uparrow$ & $E_{op}\downarrow$ & $E_{or}\downarrow$ & $E_h\downarrow$ \\
      \midrule
      Open-Loop & \srnum{34.2}\srstd{1.2} & \errnum{6.05} & \errnum{27.26} & \textbf{\errnum{2.50}} & \srnum{82.0}\srstd{0.7} & \errnum{4.33} & \errnum{24.88} & \errnum{4.84} & \srnum{91.5}\srstd{0.5} & \errnum{1.93} & \errnum{22.78} & \textbf{\errnum{1.99}} & \srnum{0.0}\srstd{0.0} & \textbf{\errnum{1.31}} & \errnum{72.62} & \textbf{\errnum{0.99}} \\
      SkillMimic & \srnum{57.0}\srstd{1.5} & \errnum{6.15} & \errnum{21.99} & \errnum{5.69} & \srnum{88.0}\srstd{0.6} & \errnum{4.56} & \errnum{17.46} & \errnum{5.70} & \srnum{87.9}\srstd{0.8} & \errnum{2.42} & \errnum{14.92} & \errnum{4.61} & \srnum{17.5}\srstd{1.0} & \errnum{8.41} & \errnum{86.51} & \errnum{36.36} \\
      DexGen-var & \srnum{4.8}\srstd{0.4} & \errnum{7.81} & \errnum{13.15} & \errnum{5.83} & \srnum{78.7}\srstd{1.0} & \errnum{8.06} & \errnum{28.63} & \errnum{5.10} & \srnum{79.0}\srstd{1.1} & \errnum{3.00} & \errnum{19.91} & \errnum{4.44} & \srnum{0.0}\srstd{0.0} & \underline{\errnum{2.23}} & \errnum{72.50} & \errnum{2.83} \\
      DexTrack-style & \textbf{\srnum{99.8}}\srstd{0.1} & \underline{\errnum{2.60}} & \textbf{\errnum{9.85}} & \underline{\errnum{2.95}} & \srnum{97.9}\srstd{0.2} & \errnum{2.35} & \errnum{10.20} & \errnum{4.80} & \textbf{\srnum{99.7}}\srstd{0.2} & \textbf{\errnum{0.95}} & \textbf{\errnum{8.75}} & \underline{\errnum{2.60}} & \srnum{34.5}\srstd{1.2} & \errnum{3.10} & \errnum{58.40} & \errnum{3.05} \\
      \midrule
      \ours w/o $r^{wrist}$ & \tablenum[table-format=2.2]{77.93}\srstd{1.1} & \textbf{\errnum{2.07}} & \errnum{14.48} & \errnum{5.55} & \srnum{93.8}\srstd{0.6} & \errnum{2.60} & \errnum{11.65} & \errnum{5.35} & \tablenum[table-format=2.2]{55.61}\srstd{0.6} & \errnum{1.20} & \errnum{12.75} & \errnum{3.04} & \srnum{42.7}\srstd{1.8} & \errnum{3.42} & \errnum{54.71} & \errnum{3.06} \\
      \ours w/o $r^{contact}$ & \srnum{0.0}\srstd{0.0} & \errnum{10.80} & \errnum{24.60} & \errnum{12.40} & \srnum{96.5}\srstd{0.6} & \errnum{2.60} & \errnum{11.30} & \errnum{5.10} & \srnum{96.0}\srstd{0.5} & \errnum{1.10} & \errnum{9.70} & \errnum{2.85} & \underline{\srnum{45.8}}\srstd{1.4} & \errnum{2.95} & \errnum{52.80} & \errnum{3.25} \\
      \ours w/o $r^{interact}$ & \srnum{72.3}\srstd{1.3} & \errnum{5.10} & \errnum{15.80} & \errnum{5.40} & \underline{\srnum{98.2}}\srstd{0.4} & \underline{\errnum{2.20}} & \underline{\errnum{10.10}} & \underline{\errnum{4.70}} & \srnum{96.1}\srstd{0.6} & \errnum{1.15} & \errnum{10.60} & \errnum{2.95} & \srnum{5.8}\srstd{0.5} & \errnum{6.80} & \errnum{71.50} & \errnum{5.20} \\
      \midrule
      \rowcolor{gray!15}
      \ours (Teacher) & \underline{\srnum{86.0}}\srstd{1.0} & \errnum{3.80} & \underline{\errnum{11.76}} & \errnum{4.32} & \textbf{\srnum{98.9}}\srstd{0.4} & \textbf{\errnum{2.01}} & \textbf{\errnum{9.83}} & \textbf{\errnum{4.50}} & \srnum{94.3}\srstd{0.4} & \errnum{1.24} & \errnum{9.88} & \errnum{3.14} & \textbf{\srnum{56.2}}\srstd{0.8} & \errnum{2.38} & \textbf{\errnum{44.96}} & \underline{\errnum{2.57}} \\
      \rowcolor{gray!15}
      \ours (Student) & \srnum{79.7}\srstd{1.2} & \errnum{4.65} & \errnum{14.80} & \errnum{5.10} & \srnum{97.7}\srstd{0.5} & \errnum{2.45} & \errnum{11.20} & \errnum{4.95} & \underline{\srnum{97.3}}\srstd{0.6} & \underline{\errnum{1.05}} & \underline{\errnum{9.40}} & \errnum{2.80} & \srnum{45.4}\srstd{1.5} & \errnum{3.20} & \underline{\errnum{52.10}} & \errnum{3.15} \\
      \midrule
      \addlinespace[0.25em]
      \multirow{2}{*}{Method} & \multicolumn{4}{c}{\textit{Throw}} & \multicolumn{4}{c}{\textit{Catch}} & \multicolumn{4}{c}{\textit{Regrasp}} & \multicolumn{4}{c}{\textbf{\textcolor{ourcolor}{Average}}} \\
      \cmidrule(lr){2-5}\cmidrule(lr){6-9}\cmidrule(lr){10-13}\cmidrule[0.6pt](lr){14-17}
        & SR (\%) $\uparrow$ & $E_{op}\downarrow$ & $E_{or}\downarrow$ & $E_h\downarrow$ & SR (\%) $\uparrow$ & $E_{op}\downarrow$ & $E_{or}\downarrow$ & $E_h\downarrow$ & SR (\%) $\uparrow$ & $E_{op}\downarrow$ & $E_{or}\downarrow$ & $E_h\downarrow$ & \cellcolor{ourcolor!10}SR (\%) $\uparrow$ & \cellcolor{ourcolor!10}$E_{op}\downarrow$ & \cellcolor{ourcolor!10}$E_{or}\downarrow$ & \cellcolor{ourcolor!10}$E_h\downarrow$ \\
      \midrule
      Open-Loop & \srnum{9.2}\srstd{0.6} & \errnum{20.20} & \errnum{91.64} & \textbf{\errnum{2.56}} & \srnum{27.9}\srstd{1.4} & \errnum{13.04} & \errnum{89.30} & \textbf{\errnum{1.51}} & \srnum{0.0}\srstd{0.0} & \errnum{21.89} & \errnum{71.84} & \textbf{\errnum{1.31}} & \cellcolor{ourcolor!10}\srnum{35.0} & \cellcolor{ourcolor!10}\errnum{9.82} & \cellcolor{ourcolor!10}\errnum{57.19} & \cellcolor{ourcolor!10}\textbf{\errnum{2.24}} \\
      SkillMimic & \srnum{0.4}\srstd{0.1} & \errnum{45.41} & \errnum{97.86} & \errnum{46.43} & \srnum{64.8}\srstd{1.3} & \errnum{6.96} & \errnum{53.65} & \errnum{4.31} & \srnum{0.0}\srstd{0.0} & \errnum{7.54} & \errnum{64.28} & \errnum{4.18} & \cellcolor{ourcolor!10}\srnum{45.1} & \cellcolor{ourcolor!10}\errnum{11.64} & \cellcolor{ourcolor!10}\errnum{50.95} & \cellcolor{ourcolor!10}\errnum{15.33} \\
      DexGen-var & \srnum{0.6}\srstd{0.2} & \errnum{41.19} & \errnum{90.00} & \errnum{44.72} & \srnum{35.9}\srstd{1.6} & \errnum{12.13} & \errnum{85.61} & \errnum{4.35} & \srnum{0.0}\srstd{0.0} & \errnum{10.47} & \errnum{58.29} & \errnum{4.92} & \cellcolor{ourcolor!10}\srnum{28.4} & \cellcolor{ourcolor!10}\errnum{12.13} & \cellcolor{ourcolor!10}\errnum{52.58} & \cellcolor{ourcolor!10}\errnum{10.31} \\
      DexTrack-style & \srnum{72.4}\srstd{1.5} & \errnum{19.80} & \errnum{47.50} & \errnum{10.20} & \textbf{\srnum{83.6}}\srstd{0.6} & \underline{\errnum{3.55}} & \underline{\errnum{46.20}} & \errnum{2.85} & \srnum{63.4}\srstd{1.0} & \errnum{2.45} & \errnum{22.80} & \errnum{3.65} & \cellcolor{ourcolor!10}\srnum{78.8} & \cellcolor{ourcolor!10}\errnum{4.97} & \cellcolor{ourcolor!10}\errnum{29.10} & \cellcolor{ourcolor!10}\errnum{4.30} \\
      \midrule
      \ours w/o $r^{wrist}$ & \srnum{82.4}\srstd{1.0} & \errnum{18.25} & \errnum{42.93} & \errnum{9.62} & \underline{\srnum{78.8}}\srstd{0.5} & \errnum{4.61} & \errnum{55.21} & \errnum{3.57} & \srnum{76.1}\srstd{1.4} & \errnum{2.03} & \underline{\errnum{16.44}} & \errnum{3.45} & \cellcolor{ourcolor!10}\srnum{72.5} & \cellcolor{ourcolor!10}\underline{\errnum{4.88}} & \cellcolor{ourcolor!10}\errnum{29.74} & \cellcolor{ourcolor!10}\errnum{4.81} \\
      \ours w/o $r^{contact}$ & \srnum{0.0}\srstd{0.0} & \errnum{45.41} & \errnum{97.86} & \errnum{46.43} & \srnum{30.7}\srstd{1.7} & \errnum{13.04} & \errnum{89.30} & \textbf{\errnum{1.51}} & \srnum{72.1}\srstd{1.1} & \errnum{2.15} & \errnum{19.60} & \errnum{3.40} & \cellcolor{ourcolor!10}\srnum{48.7} & \cellcolor{ourcolor!10}\errnum{11.15} & \cellcolor{ourcolor!10}\errnum{43.59} & \cellcolor{ourcolor!10}\errnum{10.71} \\
      \ours w/o $r^{interact}$ & \srnum{82.3}\srstd{1.2} & \errnum{17.60} & \errnum{42.80} & \errnum{9.15} & \srnum{74.1}\srstd{1.0} & \errnum{5.20} & \errnum{52.30} & \errnum{3.65} & \srnum{30.5}\srstd{1.4} & \errnum{5.60} & \errnum{39.20} & \errnum{5.70} & \cellcolor{ourcolor!10}\srnum{65.6} & \cellcolor{ourcolor!10}\errnum{6.24} & \cellcolor{ourcolor!10}\errnum{34.61} & \cellcolor{ourcolor!10}\errnum{5.25} \\
      \midrule
      \rowcolor{gray!15}
      \ours (Teacher) & \textbf{\srnum{91.9}}\srstd{1.2} & \textbf{\errnum{15.04}} & \textbf{\errnum{35.27}} & \underline{\errnum{8.34}} & \textbf{\srnum{83.6}}\srstd{0.4} & \textbf{\errnum{3.34}} & \textbf{\errnum{45.00}} & \underline{\errnum{2.67}} & \textbf{\srnum{81.9}}\srstd{0.5} & \textbf{\errnum{1.56}} & \textbf{\errnum{14.05}} & \underline{\errnum{2.90}} & \cellcolor{ourcolor!20}\textbf{\srnum{84.7}} & \cellcolor{ourcolor!20}\textbf{\errnum{4.20}} & \cellcolor{ourcolor!20}\textbf{\errnum{24.39}} & \cellcolor{ourcolor!20}\underline{\errnum{4.06}} \\
      \rowcolor{gray!15}
      \ours (Student) & \underline{\srnum{86.7}}\srstd{1.4} & \underline{\errnum{16.90}} & \underline{\errnum{40.50}} & \errnum{9.25} & \srnum{76.5}\srstd{0.9} & \errnum{4.80} & \errnum{51.60} & \errnum{3.40} & \underline{\srnum{78.5}}\srstd{1.1} & \underline{\errnum{1.95}} & \errnum{17.30} & \errnum{3.25} & \cellcolor{ourcolor!20}\underline{\srnum{80.3}} & \cellcolor{ourcolor!20}\errnum{5.00} & \cellcolor{ourcolor!20}\underline{\errnum{28.13}} & \cellcolor{ourcolor!20}\errnum{4.56} \\
      \bottomrule
    \end{tabular}
  }
\vspace{-5pt}
\end{table}

\begin{figure*}[t]
  \centering
  \begin{minipage}[t]{0.57\textwidth}
    \vspace{0pt}
    \centering
    \includegraphics[width=\linewidth]{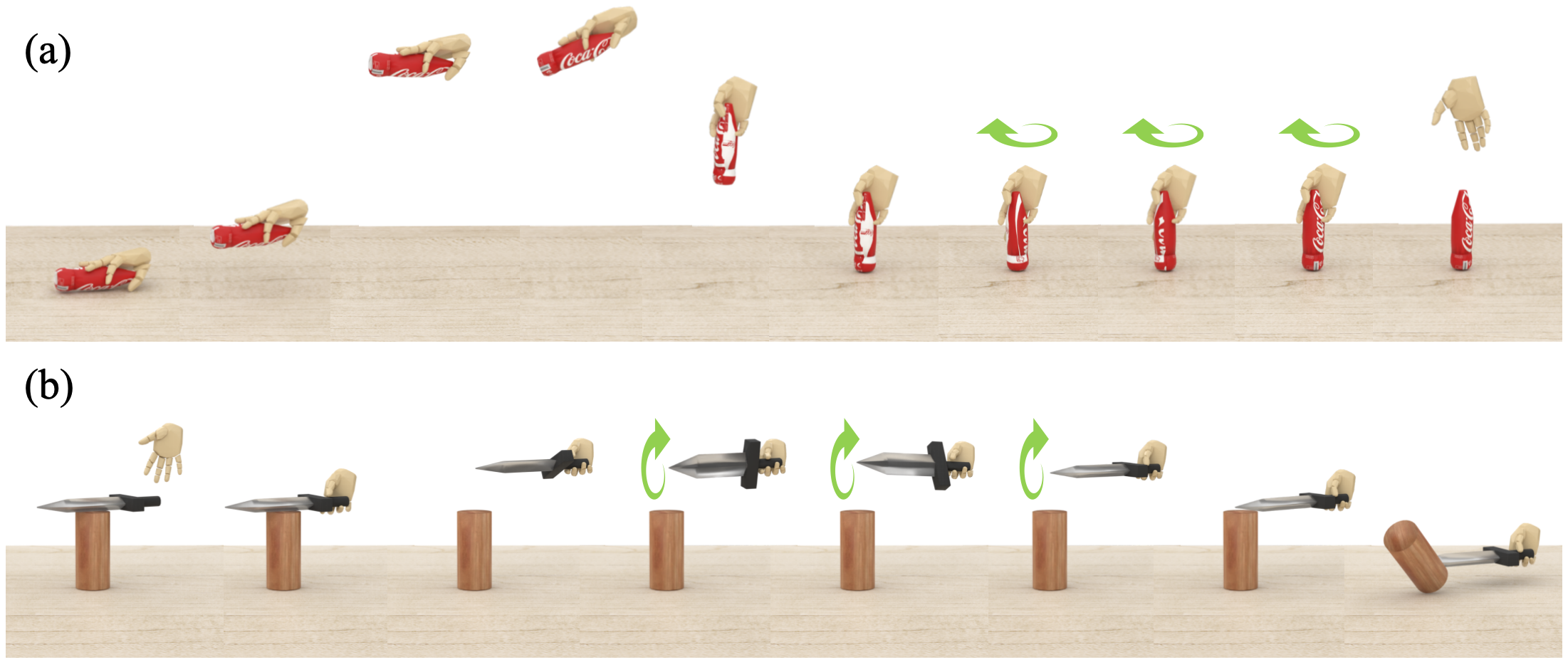}
    \captionof{figure}{Long-horizon tracking with the HOT student.}
    \label{fig: complex task}
  \end{minipage}\hfill
  \begin{minipage}[t]{0.40\textwidth}
    \vspace{0pt}
    \centering
    \captionof{table}{Zero-shot tracking of long-horizon Grasp--Move--Place sequences with the distilled HOT student.}
    \vspace{-0.1cm}
    \label{tab:hoi-long}
    {
      \renewcommand\arraystretch{1.3}
      \setlength{\tabcolsep}{8pt}
      \resizebox{\linewidth}{!}{%
        \begin{tabular}{lcccc}
          \toprule
          {Dataset} & SR (\%) $^\uparrow$ & \textbf{$E_{op}$}$^\downarrow$ & \textbf{$E_{or}$}$^\downarrow$ & \textbf{$E_{h}$}$^\downarrow$ \\
          \midrule
          Bottle & 58.54 & 9.00 & 34.77 & 4.94 \\
          Box & 84.21 & 5.69 & 34.45 & 4.69 \\
          Hammer & 57.91 & 9.40 & 38.33 & 5.37 \\
          Sword & 68.42 & 10.40 & 33.31 & 4.80 \\
          \midrule
          \textbf{Average} & \textbf{67.27} & \textbf{8.62} & \textbf{35.22} & \textbf{4.95} \\
          \bottomrule
        \end{tabular}
      }
    }
    \vspace{-0.1cm}
  \end{minipage}
\end{figure*}

\subsection{Experimental Results}
\label{sec:baseline}

We evaluate seven skills with MANO on four objects: a box, a bottle, a sword, and a hammer. For each category, we generate about 20 grasp configurations, 1000 training demonstrations per skill, and 100 test demonstrations per skill, all at 60 fps. A dedicated HOT teacher is trained for each skill-object pair.

\paragraph{Baseline configurations.}
In Tab.~\ref{tab:main}, DexTrack-style and \ours teachers use separate policies for each skill--object pair, while the \ours student is a single policy distilled across all seven skills and four objects. Additional implementation details are provided in Appendix~\ref{sec:method-details}.

\paragraph{Performance.}
Quantitative results are summarized in Tab.~\ref{tab:main}. The Open-Loop baseline performs well in Place and Move tasks, as these skills either maintain a stable force-closure grasp (Move) or involve a simple object release onto a surface (Place). However, for skills such as Grasp---where synthetic references may contain physical inconsistencies (see Fig.~\ref{fig: learn from imperfect})---the open-loop strategy often fails. In contrast, by using the same imperfect references within a reinforcement imitation learning framework, \ours yields high success rates and low tracking errors across most skills.

Both SkillMimic~\citep{wang2024skillmimic} and DexGen-var~\citep{yin2025dexteritygen} perform substantially worse than \ours. DexGen-var struggles with precise object tracking due to its lack of direct object state observation. The low performance of SkillMimic further underscores the importance of our design choices---including reward shaping, observation localization, geometry perception, residual action, and adaptive sampling---in achieving robust HOI tracking performance.

DexTrack-style performs best on Grasp and Place, while \ours teachers achieve higher success rates on Rotate, Regrasp, and Throw, yielding a higher average SR (84.7\% vs.\ 78.8\%). The gains on Rotate and Regrasp suggest greater robustness to imperfect references generated from discrete keyframes, consistent with \ours's domain randomization curriculum.

The unified HOT student outperforms all four baselines in average SR (80.3\%). Its advantages over DexTrack-style are most evident on Rotate, Throw, and Regrasp, suggesting that distillation preserves the teachers' ability to handle challenging contact transitions and imperfect references. Its modest gap to the specialist teachers (84.7\%) reflects a practical trade-off between tracking performance and policy consolidation.

\subsection{Ablation Study}
We conduct ablation studies to evaluate the impact of key design choices. Using motion clips for four objects across seven meta-skills, we analyze reward design with 1000 training and 100 test clips per skill. As shown in Tab.~\ref{tab:main}, the contact reward is particularly important for grasping and dynamic manipulation: removing it reduces Grasp and Throw success rates to 0.0\% and Catch success to 30.7\%. Tab.~\ref{tab:main} also shows that each term contributes noticeably, and the full reward yields the best overall results. 

\subsection{HOG as a Training Data Source}
\label{sec:why-hog}
We compare HOG against four representative HOI data sources: video-estimated motion, generative HOI models, motion-capture data, and demonstration-driven data augmentation. Using two objects (bowl and cylinder), we evaluate each source along two axes: \emph{data quality}, which measures physical usability for tracking, and \emph{data coverage}, which measures diversity of object poses and grasp states.

\begin{table*}[t]
  \centering
  \vspace{-5pt}
  \caption{Comparison of HOI data sources in terms of data quality and coverage.
    Penetration and jittering are reported in mm.
    GSR denotes the stable grasp rate.
    See Sec.~\ref{sec:why-hog} for details.}
  \label{tab:data_quality}
  \label{tab:data_coverage}
  \renewcommand{\arraystretch}{1.0}
  \setlength{\tabcolsep}{4pt}
  \resizebox{\linewidth}{!}{%
    \begin{tabular}{llccc|cc}
      \toprule
      \multirow{2}{*}{Source Type} & \multirow{2}{*}{Method} & \multicolumn{3}{c|}{Data Quality} & \multicolumn{2}{c}{Data Coverage} \\
      \cmidrule(lr){3-5} \cmidrule(lr){6-7}
        & & Penetration $\downarrow$ & Jittering $\downarrow$ & GSR $\uparrow$ & State-Space Coverage $\uparrow$ & Grasp Diversity $\uparrow$ \\
      \midrule
      Video Estimation & EgoGrasp & 4.52 & 2.36 & 14\% & 0.61 & 0.38 \\
      Generative Model & LatentHOI~\citep{li2025latenthoi} & 2.55 & 26.88 & 24\% & 0.08 & 0.23 \\
      Motion Capture & GRAB & 4.49 & 0.03 & 28\% & 0.62 & 0.42 \\
      Data Augmentation & DexMimicGen & 0.00 & 2.31 & 100\% & 0.02 & 0.08 \\
      \midrule
      \textbf{HOG (Ours)} & - & \textbf{1.28} & \textbf{1.43} & \textbf{100\%} & \textbf{1.97} & \textbf{0.54} \\
      \bottomrule
    \end{tabular}
  }
\end{table*}

The left half of Tab.~\ref{tab:data_quality} shows that each alternative source misses at least one key property. Video-estimated trajectories and GRAB capture human motion but exhibit substantial hand-object penetration and yield low grasp success rates after retargeting. Generative trajectories offer flexibility, yet exhibit large temporal jitter and low grasp stability. DexMimicGen maintains high trajectory quality, but only by reusing and augmenting demonstrations rather than expanding the reachable HOI state space.
The right half of Tab.~\ref{tab:data_coverage} complements this picture: HOG achieves the largest spatial coverage and the strongest grasp diversity. See Appendix~\ref{app:data-evaluation} for details of the metrics.

\vspace{-5pt}
\begin{figure*}[!t]
  \centering
  \includegraphics[width=\linewidth]{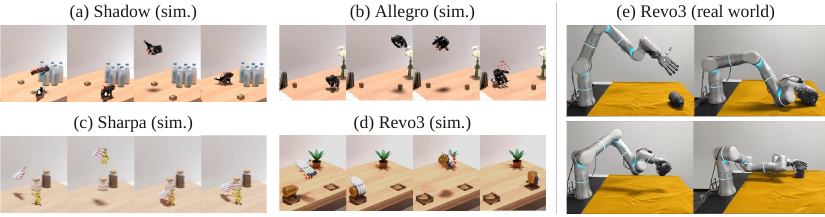}
  \vspace{-5pt}
  \caption{HOI tracking with different dexterous hands. Red dots denote hand joint tracking targets.}
  \label{fig:allegro-and-shadow}
\end{figure*}

\vspace{-5pt}
\subsection{Long-Horizon Tracking}
\label{sec:trained-compositions}
HOG synthesizes diverse long-horizon sequences at scale by concatenating meta-skill trajectories. As shown in Tab.~\ref{tab:hoi-long}, the \ours student tracks Grasp--Move--Place sequences zero-shot. For more challenging compositions involving Rotate or Regrasp, we fine-tune the student on HOG-generated composite sequences for only 1,000 epochs. The qualitative results in Fig.~\ref{fig: teaser}(b) and Fig.~\ref{fig: complex task} demonstrate tracking of Catch--Rotate--Move and Grasp--Move--Place--Rotate after this brief fine-tuning.

\vspace{-5pt}
\subsection{HOT Generalization}
\label{sec:generalization}
\paragraph{Object Generalization}
To evaluate object generalization, we distill object-specific teachers into student policies covering Grasp, Move, and Place, and compare training with 1, 5, and 15 objects. As shown in Fig.~\ref{fig:obj-generalization} and Tab.~\ref{tab:merged_policy_table}, increasing the diversity of training objects improves average tracking performance across test objects, demonstrating the benefit of multi-object training for generalization. Note that these students are trained separately from the \ours (student) evaluated in Sec.~\ref{sec:baseline}.

\vspace{-5pt}
\paragraph{Motion Transfer}
We further test this student on HOG-retargeted and refined GRAB trajectories~\citep{GRAB2020} (Appendix~\ref{sec:retarget-grab}); Fig.~\ref{fig:grabdemo} and Table~\ref{tab:grab-generalization} report transfer to these processed human-motion references. 

\vspace{-5pt}
\paragraph{Different hand morphologies.}
We evaluate the applicability of our framework to Shadow~\citep{shadow}, Allegro~\citep{allegro}, Sharpa, and Revo3 by generating HOG trajectories and training a separate \ours policy for each hand. The quantitative results in Tab.~\ref{tab:different_hands} and qualitative examples in Fig.~\ref{fig:allegro-and-shadow} show that the same data synthesis and training pipeline supports tracking across diverse hand morphologies.

\vspace{-5pt}
\begin{table*}[!t]
  \centering
  \caption{\textbf{Tracking performance across hand morphologies} on three bottle manipulation skills.}
  \vspace{-5pt}
  \label{tab:different_hands}
  \renewcommand{\arraystretch}{1.1}
  \setlength{\tabcolsep}{3pt}
  \small
  \sisetup{mode=text,detect-weight=true}
  \newcommand{\srnum}[1]{\tablenum[table-format=2.1,round-mode=places,round-precision=1,round-half=up]{#1}}
  \newcommand{\srstd}[1]{\ensuremath{^{\pm #1}}}
  \newcommand{\errnum}[1]{\tablenum[table-format=2.2]{#1}}
  \resizebox{\linewidth}{!}{%
    \begin{tabular}{l@{\hspace{8pt}}|cccc@{\hspace{8pt}}cccc@{\hspace{8pt}}cccc@{\hspace{8pt}}cccc}
      \toprule
      \multirow{2}{*}{Hand} & \multicolumn{4}{c}{\textit{Grasp}} & \multicolumn{4}{c}{\textit{Move}} & \multicolumn{4}{c}{\textit{Place}} & \multicolumn{4}{c}{\textbf{Average}} \\
      \cmidrule(lr){2-5}\cmidrule(lr){6-9}\cmidrule(lr){10-13}\cmidrule[0.6pt](lr){14-17}
        & SR (\%) $\uparrow$ & $E_{op}\downarrow$ & $E_{or}\downarrow$ & $E_h\downarrow$ & SR (\%) $\uparrow$ & $E_{op}\downarrow$ & $E_{or}\downarrow$ & $E_h\downarrow$ & SR (\%) $\uparrow$ & $E_{op}\downarrow$ & $E_{or}\downarrow$ & $E_h\downarrow$ & SR (\%) $\uparrow$ & $E_{op}\downarrow$ & $E_{or}\downarrow$ & $E_h\downarrow$ \\
      \midrule
      Shadow & \srnum{92.58}\srstd{0.6} & \errnum{1.31} & \errnum{8.39} & \errnum{3.16} & \srnum{95.82}\srstd{0.4} & \errnum{2.48} & \errnum{16.02} & \errnum{4.23} & \srnum{86.28}\srstd{0.8} & \errnum{1.15} & \errnum{6.73} & \errnum{2.85} & \srnum{91.56}\srstd{0.4} & \errnum{1.65} & \errnum{10.38} & \errnum{3.41} \\
      Allegro & \srnum{84.80}\srstd{1.0} & \errnum{2.01} & \errnum{19.50} & \errnum{6.60} & \srnum{89.78}\srstd{0.8} & \errnum{3.44} & \errnum{23.78} & \errnum{6.43} & \srnum{78.78}\srstd{1.2} & \errnum{1.66} & \errnum{13.72} & \errnum{5.23} & \srnum{84.45}\srstd{0.7} & \errnum{2.37} & \errnum{19.00} & \errnum{6.09} \\
      Sharpa & \srnum{91.24}\srstd{0.7} & \errnum{1.48} & \errnum{10.72} & \errnum{3.54} & \srnum{95.06}\srstd{0.5} & \errnum{2.62} & \errnum{17.35} & \errnum{4.51} & \srnum{85.62}\srstd{0.9} & \errnum{1.24} & \errnum{8.16} & \errnum{3.12} & \srnum{90.64}\srstd{0.5} & \errnum{1.78} & \errnum{12.08} & \errnum{3.72} \\
      Revo3 & \srnum{89.46}\srstd{0.8} & \errnum{1.62} & \errnum{12.84} & \errnum{4.05} & \srnum{93.72}\srstd{0.6} & \errnum{2.85} & \errnum{18.96} & \errnum{4.87} & \srnum{83.88}\srstd{1.0} & \errnum{1.36} & \errnum{9.72} & \errnum{3.46} & \srnum{89.02}\srstd{0.6} & \errnum{1.94} & \errnum{13.84} & \errnum{4.13} \\
      \ours Teacher & \srnum{94.44}\srstd{0.6} & \errnum{2.10} & \errnum{10.27} & \errnum{3.63} & \srnum{99.14}\srstd{0.2} & \errnum{1.55} & \errnum{7.38} & \errnum{3.62} & \srnum{98.18}\srstd{0.3} & \errnum{1.04} & \errnum{9.05} & \errnum{2.65} & \srnum{97.25}\srstd{0.3} & \errnum{1.56} & \errnum{8.90} & \errnum{3.30} \\
      \bottomrule
    \end{tabular}
  }
\vspace{-5pt}
\end{table*}

\vspace{-4pt}\section{Conclusion}
We presented a system for generalizable dexterous manipulation that combines synthetic data generation with tracking-based control. Our approach consists of HOG, a scalable framework for synthesizing hand-object trajectories across diverse meta-skills, and HOT, a tracking policy trained via reinforcement imitation learning on such data. The resulting controller demonstrates effective generalization to unseen objects, long-horizon tasks, and even real-world human-object motion data. We further validate the framework across different hand morphologies and demonstrate cross-simulator transfer. By functioning effectively without human demonstrations while remaining compatible with them, our method offers a flexible, scalable, and powerful foundation for future work on general-purpose dexterous manipulation.

\begin{ack}
This work was supported by the Research Grants Council of HKSAR (Grant No. AoE/E-601/24-N), the Shenzhen Loop Area Institute (Grant No. 2026NSFP006), and the Hubei Provincial Science and Technology Plan Project (Grant No. 2025BEA002).

The authors declare no competing interests.
\end{ack}

\bibliographystyle{plainnat}
\bibliography{main}

@String(CVPR= {IEEE Conf. Comput. Vis. Pattern Recog.})

@String(ECCV= {Eur. Conf. Comput. Vis.})

@String(CVPR  = {CVPR})

@String(ECCV  = {ECCV})

@inproceedings{zhang2024artigrasp,
  title={ArtiGrasp: Physically plausible synthesis of bi-manual dexterous grasping and articulation},
  author={Zhang, Hui and Christen, Sammy and Fan, Zicong and Zheng, Luocheng and Hwangbo, Jemin and Song, Jie and Hilliges, Otmar},
  booktitle={2024 International Conference on 3D Vision (3DV)},
  pages={235--246},
  year={2024},
  organization={IEEE}
}

@article{fu2026egograsp,
  title={EgoGrasp: World-Space Hand-Object Interaction Estimation from Egocentric Videos},
  author={Fu, Hongming and Wang, Wenjia and Qiao, Xiaozhen and Potamias, Rolandos Alexandros and Komura, Taku and Yang, Shuo and Liu, Zheng and Zhao, Bo},
  journal={arXiv preprint arXiv:2601.01050},
  year={2026}
}

@inproceedings{GRAB2020,
  title = {{GRAB}: A Dataset of Whole-Body Human Grasping of Objects},
  author = {Taheri, Omid and Ghorbani, Nima and Black, Michael J. and Tzionas, Dimitrios},
  booktitle = {European Conference on Computer Vision (ECCV)},
  year = {2020},
  url = {https://grab.is.tue.mpg.de}
}

@article{garrett2024skillmimicgen,
  title={SkillMimicGen: Automated Demonstration Generation for Efficient Skill Learning and Deployment},
  author={Garrett, Caelan and Mandlekar, Ajay and Wen, Bowen and Fox, Dieter},
  journal={arXiv preprint arXiv:2410.18907},
  year={2024}
}

@article{she2022learning,
  title={Learning high-DOF reaching-and-grasping via dynamic representation of gripper-object interaction},
  author={She, Qijin and Hu, Ruizhen and Xu, Juzhan and Liu, Min and Xu, Kai and Huang, Hui},
  journal={arXiv preprint arXiv:2204.13998},
  year={2022}
}

@inproceedings{she2024learning,
  title={Learning Cross-Hand Policies of High-DOF Reaching and Grasping},
  author={She, Qijin and Zhang, Shishun and Ye, Yunfan and Hu, Ruizhen and Xu, Kai},
  booktitle={European Conference on Computer Vision},
  pages={269--285},
  year={2024},
  organization={Springer}
}

@inproceedings{li2025maniptrans,
  title={Maniptrans: Efficient dexterous bimanual manipulation transfer via residual learning},
  author={Li, Kailin and Li, Puhao and Liu, Tengyu and Li, Yuyang and Huang, Siyuan},
  booktitle={Proceedings of the Computer Vision and Pattern Recognition Conference},
  pages={6991--7003},
  year={2025}
}

@article{achiam2023gpt,
  title={Gpt-4 technical report},
  author={Achiam, Josh and Adler, Steven and Agarwal, Sandhini and Ahmad, Lama and Akkaya, Ilge and Aleman, Florencia Leoni and Almeida, Diogo and Altenschmidt, Janko and Altman, Sam and Anadkat, Shyamal and others},
  journal={arXiv preprint arXiv:2303.08774},
  year={2023}
}

@article{team2023gemini,
  title={Gemini: a family of highly capable multimodal models},
  author={Team, Gemini and Anil, Rohan and Borgeaud, Sebastian and Alayrac, Jean-Baptiste and Yu, Jiahui and Soricut, Radu and Schalkwyk, Johan and Dai, Andrew M and Hauth, Anja and Millican, Katie and others},
  journal={arXiv preprint arXiv:2312.11805},
  year={2023}
}

@inproceedings{cha2024text2hoi,
  title={Text2hoi: Text-guided 3d motion generation for hand-object interaction},
  author={Cha, Junuk and Kim, Jihyeon and Yoon, Jae Shin and Baek, Seungryul},
  booktitle={Proceedings of the IEEE/CVF Conference on Computer Vision and Pattern Recognition},
  pages={1577--1585},
  year={2024}
}

@inproceedings{cao2021reconstructing,
  title={Reconstructing hand-object interactions in the wild},
  author={Cao, Zhe and Radosavovic, Ilija and Kanazawa, Angjoo and Malik, Jitendra},
  booktitle={Proceedings of the IEEE/CVF International Conference on Computer Vision},
  pages={12417--12426},
  year={2021}
}

@inproceedings{wen2023bundlesdf,
  title={Bundlesdf: Neural 6-dof tracking and 3d reconstruction of unknown objects},
  author={Wen, Bowen and Tremblay, Jonathan and Blukis, Valts and Tyree, Stephen and M{\"u}ller, Thomas and Evans, Alex and Fox, Dieter and Kautz, Jan and Birchfield, Stan},
  booktitle={Proceedings of the IEEE/CVF Conference on Computer Vision and Pattern Recognition},
  pages={606--617},
  year={2023}
}

@article{chen2024object,
  title={Object-centric dexterous manipulation from human motion data},
  author={Chen, Yuanpei and Wang, Chen and Yang, Yaodong and Liu, C Karen},
  journal={arXiv preprint arXiv:2411.04005},
  year={2024}
}

@article{yin2023rotating,
  title={Rotating without seeing: Towards in-hand dexterity through touch},
  author={Yin, Zhao-Heng and Huang, Binghao and Qin, Yuzhe and Chen, Qifeng and Wang, Xiaolong},
  journal={arXiv preprint arXiv:2303.10880},
  year={2023}
}

@article{liu2025dextrack,
  title={DexTrack: Towards Generalizable Neural Tracking Control for Dexterous Manipulation from Human References},
  author={Liu, Xueyi and Adalibieke, Jianibieke and Han, Qianwei and Qin, Yuzhe and Yi, Li},
  journal={arXiv preprint arXiv:2502.09614},
  year={2025}
}

@article{liu2021synthesizing,
  title={Synthesizing diverse and physically stable grasps with arbitrary hand structures using differentiable force closure estimator},
  author={Liu, Tengyu and Liu, Zeyu and Jiao, Ziyuan and Zhu, Yixin and Zhu, Song-Chun},
  journal={IEEE Robotics and Automation Letters},
  volume={7},
  number={1},
  pages={470--477},
  year={2021},
  publisher={IEEE}
}

@article{miller2004graspit,
  title={Graspit! a versatile simulator for robotic grasping},
  author={Miller, Andrew T and Allen, Peter K},
  journal={IEEE Robotics \& Automation Magazine},
  volume={11},
  number={4},
  pages={110--122},
  year={2004},
  publisher={IEEE}
}

@inproceedings{hasson2019learning,
  title={Learning joint reconstruction of hands and manipulated objects},
  author={Hasson, Yana and Varol, Gul and Tzionas, Dimitrios and Kalevatykh, Igor and Black, Michael J and Laptev, Ivan and Schmid, Cordelia},
  booktitle={Proceedings of the IEEE/CVF conference on computer vision and pattern recognition},
  pages={11807--11816},
  year={2019}
}

@inproceedings{chao2021dexycb,
  title={DexYCB: A benchmark for capturing hand grasping of objects},
  author={Chao, Yu-Wei and Yang, Wei and Xiang, Yu and Molchanov, Pavlo and Handa, Ankur and Tremblay, Jonathan and Narang, Yashraj S and Van Wyk, Karl and Iqbal, Umar and Birchfield, Stan and others},
  booktitle={Proceedings of the IEEE/CVF Conference on Computer Vision and Pattern Recognition},
  pages={9044--9053},
  year={2021}
}

@article{yin2025dexteritygen,
  title={DexterityGen: Foundation Controller for Unprecedented Dexterity},
  author={Yin, Zhao-Heng and Wang, Changhao and Pineda, Luis and Hogan, Francois and Bodduluri, Krishna and Sharma, Akash and Lancaster, Patrick and Prasad, Ishita and Kalakrishnan, Mrinal and Malik, Jitendra and others},
  journal={arXiv preprint arXiv:2502.04307},
  year={2025}
}

@article{yu2025skillmimic,
  title={SkillMimic-V2: Learning Robust and Generalizable Interaction Skills from Sparse and Noisy Demonstrations},
  author={Yu, Runyi and Wang, Yinhuai and Zhao, Qihan and Tsui, Hok Wai and Wang, Jingbo and Tan, Ping and Chen, Qifeng},
  journal={arXiv preprint arXiv:2505.02094},
  year={2025}
}

@inproceedings{wang2023dexgraspnet,
  title={Dexgraspnet: A large-scale robotic dexterous grasp dataset for general objects based on simulation},
  author={Wang, Ruicheng and Zhang, Jialiang and Chen, Jiayi and Xu, Yinzhen and Li, Puhao and Liu, Tengyu and Wang, He},
  booktitle={2023 IEEE International Conference on Robotics and Automation (ICRA)},
  pages={11359--11366},
  year={2023},
  organization={IEEE}
}

@article{chen2023sequential,
  title={Sequential dexterity: Chaining dexterous policies for long-horizon manipulation},
  author={Chen, Yuanpei and Wang, Chen and Fei-Fei, Li and Liu, C Karen},
  journal={arXiv preprint arXiv:2309.00987},
  year={2023}
}

@article{huang2023dynamic,
  title={Dynamic handover: Throw and catch with bimanual hands},
  author={Huang, Binghao and Chen, Yuanpei and Wang, Tianyu and Qin, Yuzhe and Yang, Yaodong and Atanasov, Nikolay and Wang, Xiaolong},
  journal={arXiv preprint arXiv:2309.05655},
  year={2023}
}

@article{chen2023bi,
  title={Bi-dexhands: Towards human-level bimanual dexterous manipulation},
  author={Chen, Yuanpei and Geng, Yiran and Zhong, Fangwei and Ji, Jiaming and Jiang, Jiechuang and Lu, Zongqing and Dong, Hao and Yang, Yaodong},
  journal={IEEE Transactions on Pattern Analysis and Machine Intelligence},
  year={2023},
  publisher={IEEE}
}

@article{pumacay2024colosseum,
  title={The colosseum: A benchmark for evaluating generalization for robotic manipulation},
  author={Pumacay, Wilbert and Singh, Ishika and Duan, Jiafei and Krishna, Ranjay and Thomason, Jesse and Fox, Dieter},
  journal={arXiv preprint arXiv:2402.08191},
  year={2024}
}

@inproceedings{tobin2017domain,
  title={Domain randomization for transferring deep neural networks from simulation to the real world},
  author={Tobin, Josh and Fong, Rachel and Ray, Alex and Schneider, Jonas and Zaremba, Wojciech and Abbeel, Pieter},
  booktitle={2017 IEEE/RSJ international conference on intelligent robots and systems (IROS)},
  pages={23--30},
  year={2017},
  organization={IEEE}
}

@inproceedings{peng2018sim,
  title={Sim-to-real transfer of robotic control with dynamics randomization},
  author={Peng, Xue Bin and Andrychowicz, Marcin and Zaremba, Wojciech and Abbeel, Pieter},
  booktitle={2018 IEEE international conference on robotics and automation (ICRA)},
  pages={3803--3810},
  year={2018},
  organization={IEEE}
}

@article{gao2024efficient,
  title={Efficient Data Collection for Robotic Manipulation via Compositional Generalization},
  author={Gao, Jensen and Xie, Annie and Xiao, Ted and Finn, Chelsea and Sadigh, Dorsa},
  journal={arXiv preprint arXiv:2403.05110},
  year={2024}
}

@article{jiang2024dexmimicgen,
  title={DexMimicGen: Automated Data Generation for Bimanual Dexterous Manipulation via Imitation Learning},
  author={Jiang, Zhenyu and Xie, Yuqi and Lin, Kevin and Xu, Zhenjia and Wan, Weikang and Mandlekar, Ajay and Fan, Linxi and Zhu, Yuke},
  journal={arXiv preprint arXiv:2410.24185},
  year={2024}
}

@article{mandlekar2023mimicgen,
  title={Mimicgen: A data generation system for scalable robot learning using human demonstrations},
  author={Mandlekar, Ajay and Nasiriany, Soroush and Wen, Bowen and Akinola, Iretiayo and Narang, Yashraj and Fan, Linxi and Zhu, Yuke and Fox, Dieter},
  journal={arXiv preprint arXiv:2310.17596},
  year={2023}
}

@article{yang2020multi,
  title={Multi-task reinforcement learning with soft modularization},
  author={Yang, Ruihan and Xu, Huazhe and Wu, Yi and Wang, Xiaolong},
  journal={Advances in Neural Information Processing Systems},
  volume={33},
  pages={4767--4777},
  year={2020}
}

@article{rajeswaran2018learning,
  title={Learning Complex Dexterous Manipulation with Deep Reinforcement Learning and Demonstrations},
  author={Rajeswaran, Aravind and Kumar, Vikash and Gupta, Abhishek and Vezzani, Giulia and Schulman, John and Todorov, Emanuel and Levine, Sergey},
  journal={Robotics: Science and Systems XIV},
  year={2018},
  publisher={Robotics: Science and Systems Foundation}
}

@inproceedings{wang2024skillmimic,
  title={Skillmimic: Learning reusable basketball skills from demonstrations},
  author={Wang, Yinhuai and Zhao, Qihan and Yu, Runyi and Tsui, Hok Wai and Zeng, Ailing and Lin, Jing and Luo, Zhengyi and Yu, Jiwen and Li, Xiu and Chen, Qifeng and others},
  booktitle={Proceedings of the IEEE/CVF Conference on Computer Vision and Pattern Recognition},
  year={2025}
}

@inproceedings{todorov2012mujoco,
  title={Mujoco: A physics engine for model-based control},
  author={Todorov, Emanuel and Erez, Tom and Tassa, Yuval},
  booktitle={2012 IEEE/RSJ international conference on intelligent robots and systems},
  pages={5026--5033},
  year={2012},
  organization={IEEE}
}

@inproceedings{allshire2022transferring,
  title={Transferring dexterous manipulation from gpu simulation to a remote real-world trifinger},
  author={Allshire, Arthur and MittaI, Mayank and Lodaya, Varun and Makoviychuk, Viktor and Makoviichuk, Denys and Widmaier, Felix and W{\"u}thrich, Manuel and Bauer, Stefan and Handa, Ankur and Garg, Animesh},
  booktitle={2022 IEEE/RSJ International Conference on Intelligent Robots and Systems (IROS)},
  pages={11802--11809},
  year={2022},
  organization={IEEE}
}

@article{andrychowicz2020learning,
  title={Learning dexterous in-hand manipulation},
  author={Andrychowicz, OpenAI: Marcin and Baker, Bowen and Chociej, Maciek and Jozefowicz, Rafal and McGrew, Bob and Pachocki, Jakub and Petron, Arthur and Plappert, Matthias and Powell, Glenn and Ray, Alex and others},
  journal={The International Journal of Robotics Research},
  volume={39},
  number={1},
  pages={3--20},
  year={2020},
  publisher={SAGE Publications Sage UK: London, England}
}

@article{fang2023anygrasp,
  title={Anygrasp: Robust and efficient grasp perception in spatial and temporal domains},
  author={Fang, Hao-Shu and Wang, Chenxi and Fang, Hongjie and Gou, Minghao and Liu, Jirong and Yan, Hengxu and Liu, Wenhai and Xie, Yichen and Lu, Cewu},
  journal={IEEE Transactions on Robotics},
  year={2023},
  publisher={IEEE}
}

@inproceedings{wan2023unidexgrasp++,
  title={Unidexgrasp++: Improving dexterous grasping policy learning via geometry-aware curriculum and iterative generalist-specialist learning},
  author={Wan, Weikang and Geng, Haoran and Liu, Yun and Shan, Zikang and Yang, Yaodong and Yi, Li and Wang, He},
  booktitle={Proceedings of the IEEE/CVF International Conference on Computer Vision},
  pages={3891--3902},
  year={2023}
}

@inproceedings{xu2023unidexgrasp,
  title={Unidexgrasp: Universal robotic dexterous grasping via learning diverse proposal generation and goal-conditioned policy},
  author={Xu, Yinzhen and Wan, Weikang and Zhang, Jialiang and Liu, Haoran and Shan, Zikang and Shen, Hao and Wang, Ruicheng and Geng, Haoran and Weng, Yijia and Chen, Jiayi and others},
  booktitle={Proceedings of the IEEE/CVF Conference on Computer Vision and Pattern Recognition},
  pages={4737--4746},
  year={2023}
}

@inproceedings{zhang2024graspxl,
  title={Graspxl: Generating grasping motions for diverse objects at scale},
  author={Zhang, Hui and Christen, Sammy and Fan, Zicong and Hilliges, Otmar and Song, Jie},
  booktitle={European Conference on Computer Vision},
  pages={386--403},
  year={2024},
  organization={Springer}
}

@article{shadow,
  title={Shadow Robot},
  author={Shadow Robot Developers},
  journal={https://www.shadowrobot.com/dexterous-hand-series/},
  year={2005}
}

@article{romero2022embodied,
  title={Embodied hands: Modeling and capturing hands and bodies together},
  author={Romero, Javier and Tzionas, Dimitrios and Black, Michael J},
  journal={arXiv preprint arXiv:2201.02610},
  year={2022}
}

@article{wang2023physhoi,
  title={Physhoi: Physics-based imitation of dynamic human-object interaction},
  author={Wang, Yinhuai and Lin, Jing and Zeng, Ailing and Luo, Zhengyi and Zhang, Jian and Zhang, Lei},
  journal={arXiv preprint arXiv:2312.04393},
  year={2023}
}

@article{fu2024humanplus,
  title={HumanPlus: Humanoid Shadowing and Imitation from Humans},
  author={Fu, Zipeng and Zhao, Qingqing and Wu, Qi and Wetzstein, Gordon and Finn, Chelsea},
  journal={arXiv preprint arXiv:2406.10454},
  year={2024}
}

@article{cheng2024open,
  title={Open-television: Teleoperation with immersive active visual feedback},
  author={Cheng, Xuxin and Li, Jialong and Yang, Shiqi and Yang, Ge and Wang, Xiaolong},
  journal={arXiv preprint arXiv:2407.01512},
  year={2024}
}

@article{he2024omnih2o,
  title={OmniH2O: Universal and Dexterous Human-to-Humanoid Whole-Body Teleoperation and Learning},
  author={He, Tairan and Luo, Zhengyi and He, Xialin and Xiao, Wenli and Zhang, Chong and Zhang, Weinan and Kitani, Kris and Liu, Changliu and Shi, Guanya},
  journal={arXiv preprint arXiv:2406.08858},
  year={2024}
}

@article{banerjee2024introducing,
  title={Introducing HOT3D: An Egocentric Dataset for 3D Hand and Object Tracking},
  author={Banerjee, Prithviraj and Shkodrani, Sindi and Moulon, Pierre and Hampali, Shreyas and Zhang, Fan and Fountain, Jade and Miller, Edward and Basol, Selen and Newcombe, Richard and Wang, Robert and others},
  journal={arXiv preprint arXiv:2406.09598},
  year={2024}
}

@article{schulman2017proximal,
  title={Proximal policy optimization algorithms},
  author={Schulman, John and Wolski, Filip and Dhariwal, Prafulla and Radford, Alec and Klimov, Oleg},
  journal={arXiv preprint arXiv:1707.06347},
  year={2017}
}

@inproceedings{fan2023arctic,
  title={ARCTIC: A Dataset for Dexterous Bimanual Hand-Object Manipulation},
  author={Fan, Zicong and Taheri, Omid and Tzionas, Dimitrios and Kocabas, Muhammed and Kaufmann, Manuel and Black, Michael J and Hilliges, Otmar},
  booktitle={Proceedings of the IEEE/CVF Conference on Computer Vision and Pattern Recognition},
  pages={12943--12954},
  year={2023}
}

@inproceedings{luo2023perpetual,
  title={Perpetual humanoid control for real-time simulated avatars},
  author={Luo, Zhengyi and Cao, Jinkun and Kitani, Kris and Xu, Weipeng and others},
  booktitle={Proceedings of the IEEE/CVF International Conference on Computer Vision},
  pages={10895--10904},
  year={2023}
}

@inproceedings{
makoviychuk2021isaac,
title={Isaac Gym: High Performance {GPU} Based Physics Simulation For Robot Learning},
author={Viktor Makoviychuk and Lukasz Wawrzyniak and Yunrong Guo and Michelle Lu and Kier Storey and Miles Macklin and David Hoeller and Nikita Rudin and Arthur Allshire and Ankur Handa and Gavriel State},
booktitle={Thirty-fifth Conference on Neural Information Processing Systems Datasets and Benchmarks Track (Round 2)},
year={2021},
url={https://openreview.net/forum?id=fgFBtYgJQX_}
}

@article{xu2025intermimic,
  title={Intermimic: Towards universal whole-body control for physics-based human-object interactions},
  author={Xu, Sirui and Ling, Hung Yu and Wang, Yu-Xiong and Gui, Liang-Yan},
  journal={arXiv preprint arXiv:2502.20390},
  year={2025}
}

@inproceedings{juravsky2024superpadl,
  title={Superpadl: Scaling language-directed physics-based control with progressive supervised distillation},
  author={Juravsky, Jordan and Guo, Yunrong and Fidler, Sanja and Peng, Xue Bin},
  booktitle={ACM SIGGRAPH 2024 Conference Papers},
  pages={1--11},
  year={2024}
}

@article{allegro,
  title={Allegro Hand},
  author={Wonik Robotics},
  journal={https://www.allegrohand.com/},
  year={2013}
}

@inproceedings{li2025latenthoi,
  title={Latenthoi: On the generalizable hand object motion generation with latent hand diffusion},
  author={Li, Muchen and Christen, Sammy and Wan, Chengde and Cai, Yujun and Liao, Renjie and Sigal, Leonid and Ma, Shugao},
  booktitle={2025 IEEE/CVF Conference on Computer Vision and Pattern Recognition (CVPR)},
  pages={17416--17425},
  year={2025},
  organization={IEEE}
}

\clearpage
\appendix
\section{Additional Experiments}
\begingroup
\setlength{\textfloatsep}{8pt plus 2pt minus 2pt}
\setlength{\floatsep}{6pt plus 2pt minus 2pt}
\setlength{\intextsep}{6pt plus 2pt minus 2pt}
\makeatletter
\setlength{\@fptop}{0pt}
\setlength{\@fpsep}{8pt}
\setlength{\@fpbot}{0pt plus 1fil}
\makeatother

\subsection{Impact of Data Volume and Model Design}
\label{sec:additional-ablation}
We investigate the impact of three key design choices in our approach through comprehensive ablation studies. First, we examine critical \ours design elements, specifically the localization and residualization of input-output representations that simplify the learning distribution (\ours w/o LR). Second, we study the effectiveness of our data augmentation strategies (\ours w/o DA). Third, we analyze how model performance scales with training data quantity.

The experiments are conducted on three fundamental meta-skills---Grasp, Move, and Place---with each skill trained using 10, 100, or 1000 demonstrations. Tab.~\ref{tab:ablation} presents the quantitative results of our ablation studies. The localization and residualization of the input-output representations of \ours prove crucial for generalization, allowing the model to achieve superior performance even with limited training data. This validates our hypothesis that simplifying the input-output distribution enhances learning efficiency. Data augmentation techniques consistently boost generalization performance across all skills, while increased training data leads to proportional performance improvements, demonstrating our method's ability to effectively utilize additional demonstrations.

\begin{table*}[h]
  \caption{Ablation study of model design and data volume.}
  \centering
  \resizebox{1\linewidth}{!}{%
    \begin{tabular}{l|ccc|ccc|ccc}
      \toprule
      \multirow{2}{*}{Method}&
      \multicolumn{3}{c}{\textit{10 Clips}} &
      \multicolumn{3}{c}{\textit{100 Clips}} &
      \multicolumn{3}{c}{\textit{1000 Clips}}\\

        & \multicolumn{1}{c}{\textit{\ours}} &
      \multicolumn{1}{c}{\textit{\ours w/o LR}} &
      \multicolumn{1}{c}{\textit{\ours w/o DA}} &
      \multicolumn{1}{c}{\textit{\ours}} &
      \multicolumn{1}{c}{\textit{\ours w/o LR}} &
      \multicolumn{1}{c}{\textit{\ours w/o DA}} &\multicolumn{1}{c}{\textit{\ours}} &
      \multicolumn{1}{c}{\textit{\ours w/o LR}} &
      \multicolumn{1}{c}{\textit{\ours w/o DA}}
      \\
      \addlinespace[0.15em]
      \hline
      \addlinespace[0.15em]
      Grasp  & 31.49\% & 19.41\% & 31.60\% & 73.24\% & 71.38\% & 60.22\% & 94.44\% & 64.97\% & 70.79\%\\

      Move & 79.85\% & 62.04\% & 79.75\% & 97.67\% & 94.66\% & 90.89\% & 99.14\% & 95.18\% & 95.20\% \\

      Place & 84.73\% & 28.12\% & 84.99\% & 100\% & 92.29\% & 100\% & 98.18\% & 100\% & 99.63\%\\
      \addlinespace[0.15em]
      \hline
      \addlinespace[0.15em]
      Avg. & 65.36\% & 36.52\% & 65.45\%
        & 90.30\% & 86.11\% & 83.70\% & 97.25\% &
      86.72\% & 88.54\% \\

      \bottomrule
    \end{tabular}
  }
  \label{tab:ablation}
\end{table*}

\subsection{Refining HOI Data Through Tracking}

\label{sec:penetration}
To quantitatively demonstrate \ours's ability to refine imperfect reference data through tracking, we evaluate the trained \ours on bottle Grasp sequences from the test set generated by HOG, using object-hand penetration as the key metric. As shown in Tab.~\ref{tab:penetration}, the original reference data contain noticeable penetration, while the trajectories produced by \ours exhibit significantly reduced penetration, validating its capacity to correct physically implausible artifacts. Figure~\ref{fig: learn from imperfect} provides qualitative examples of refining imperfect synthetic demonstrations for sword grasp and bottle regrasp.

\begin{table}[htbp]
  \centering
  \caption{Hand--object penetration before and after HOT tracking. Penetration depths are reported in mm; lower is better.}
  \label{tab:penetration}
  \begin{tabular}{lcc}
    \toprule
    \textbf{Trajectory} & \textbf{Mean Depth} $\downarrow$ & \textbf{Max Depth} $\downarrow$\\
    \midrule
    HOG reference & $2.3 \pm 1.3$ & $9.2 \pm 7.1$ \\
    HOT rollout & $\mathbf{1.2 \pm 0.8}$ & $\mathbf{2.8 \pm 1.3}$\\
    \bottomrule
  \end{tabular}
\end{table}

\begin{figure}[htbp]
  \centering
  \includegraphics[width=0.8\linewidth]{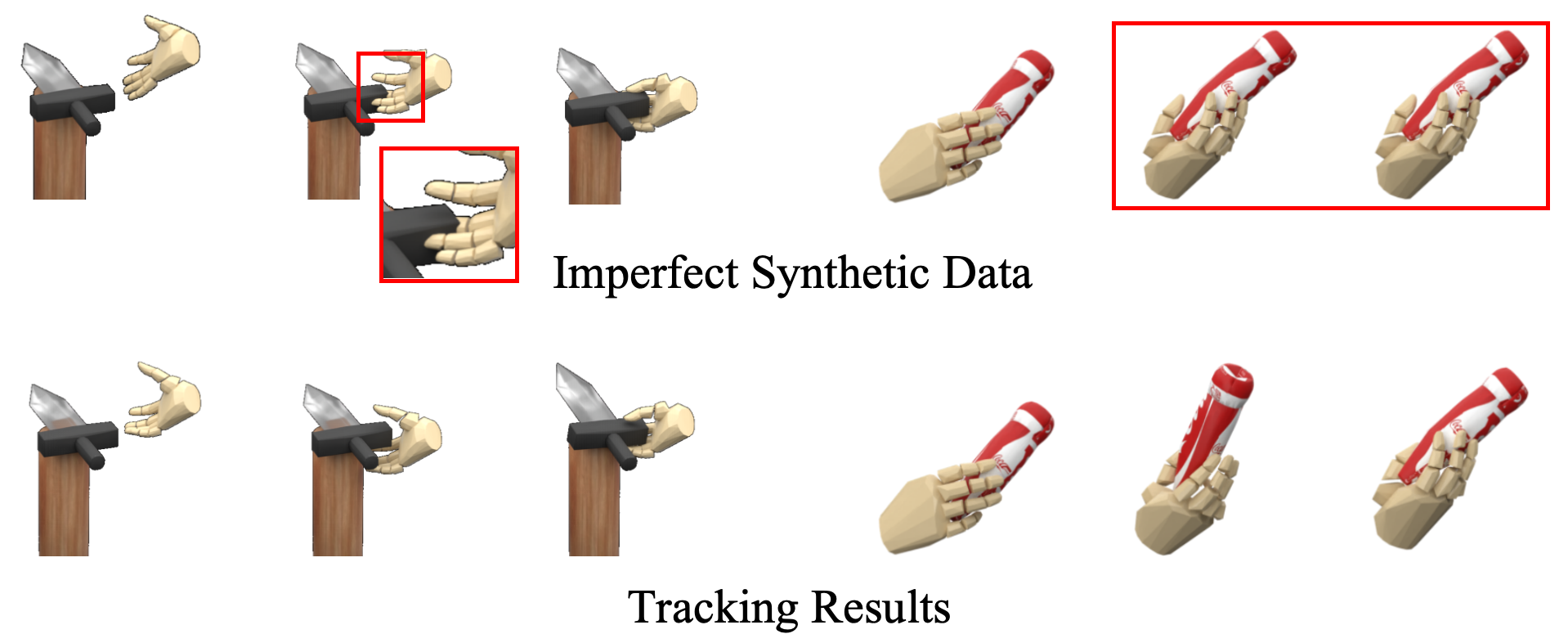}
  \caption{\ours is able to track and refine imperfect synthetic HOI demonstrations.}
  \label{fig: learn from imperfect}
\end{figure}

%

\begin{figure*}[!t]
  \centering
  \includegraphics[width=0.95\linewidth]{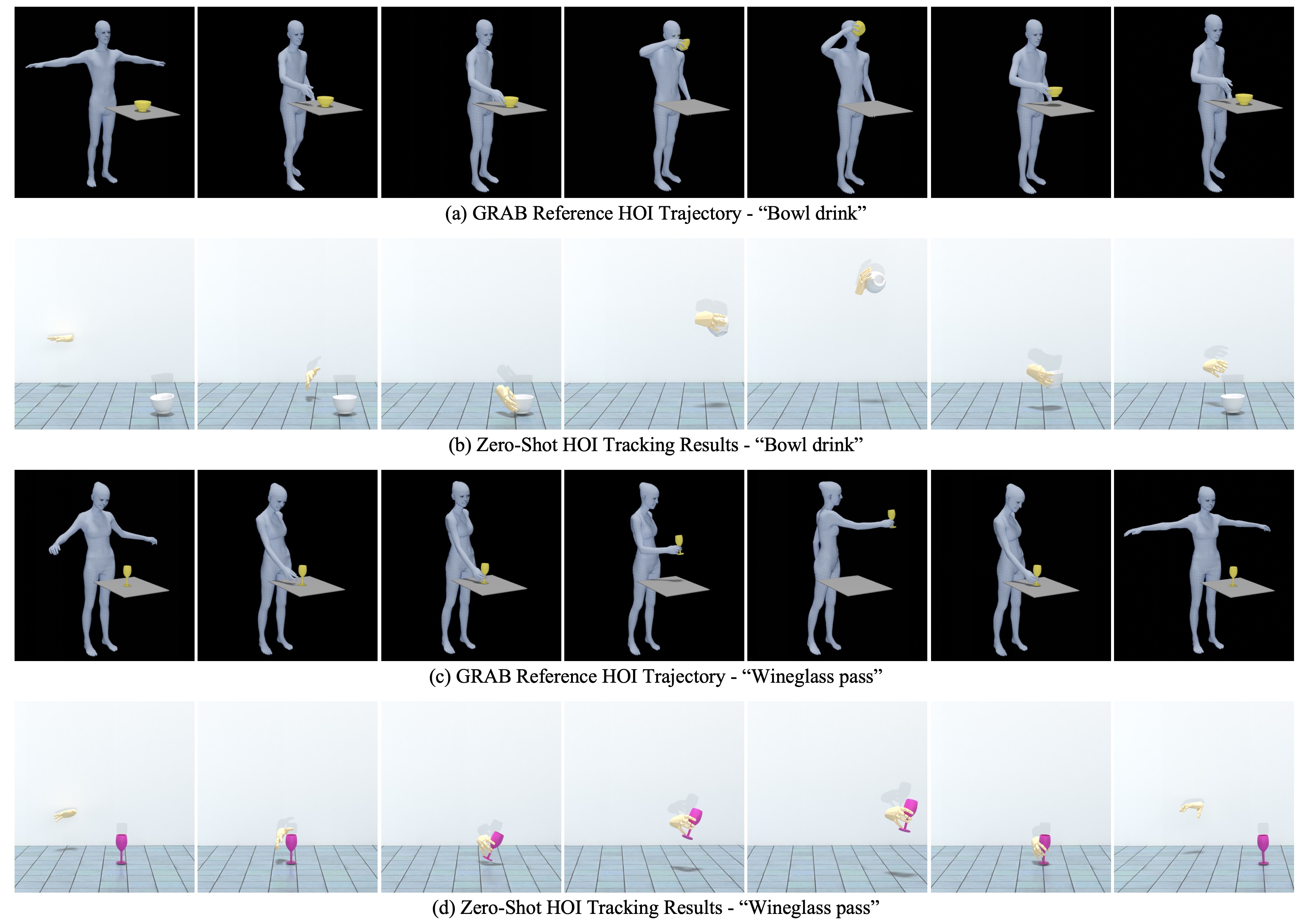}
  \caption{Zero-shot HOI tracking on GRAB data.}

  \label{fig:grabdemo}
\end{figure*}

\begin{table*}[!t]
  \centering
  \caption{HOI tracking performance on GRAB. $\mathrm{SR}_{\mathrm{grasp}}$ and $\mathrm{SR}_{\mathrm{traj}}$ report successful trials / total trials for stable grasping and full-trajectory tracking, respectively. Overall sums trial counts and averages errors.}
  \label{tab:grab-generalization}
  {\fontsize{10}{10}\selectfont
    \renewcommand\arraystretch{1.1}
    \setlength{\tabcolsep}{6pt}
    \resizebox{0.7\linewidth}{!}{%
      \begin{tabular}{lccccc}
        \toprule
        Dataset & $\mathrm{SR}_{\mathrm{grasp}} \uparrow$ & $\mathrm{SR}_{\mathrm{traj}} \uparrow$ & $E_{op} \downarrow$ & $E_{or} \downarrow$ & $E_{k} \downarrow$ \\
        \midrule
        GRAB-Apple & 3/3 & 3/3 & 1.64 & 46.43 & 4.82 \\
        GRAB-Bowl & 3/3 & 3/3 & 2.59 & 49.22 & 5.03 \\
        GRAB-Cup & 3/3 & 1/3 & 2.58 & 15.48 & 9.56 \\
        GRAB-Hammer & 2/3 & 2/3 & 5.89 & 36.36 & 5.75 \\
        GRAB-Stanfordbunny & 1/3 & 1/3 & 9.51 & 68.07 & 5.02 \\
        GRAB-Stapler & 2/3 & 2/3 & 4.26 & 29.59 & 5.62 \\
        GRAB-Train & 3/3 & 2/3 & 12.81 & 42.74 & 6.47 \\
        GRAB-Torus\_large & 1/2 & 0/2 & 24.77 & 89.07 & 11.71 \\
        GRAB-Waterbottle & 3/4 & 3/4 & 4.38 & 19.40 & 4.33 \\
        GRAB-Wineglass & 2/4 & 1/4 & 38.13 & 44.55 & 6.10 \\
        \midrule
        \textbf{Overall} & \textbf{23/31} & \textbf{18/31} & \textbf{10.66} & \textbf{44.09} & \textbf{6.44} \\
        \bottomrule
      \end{tabular}
    } }
\end{table*}

\begin{figure*}[!t]
  \centering
  \includegraphics[width=0.95\linewidth]{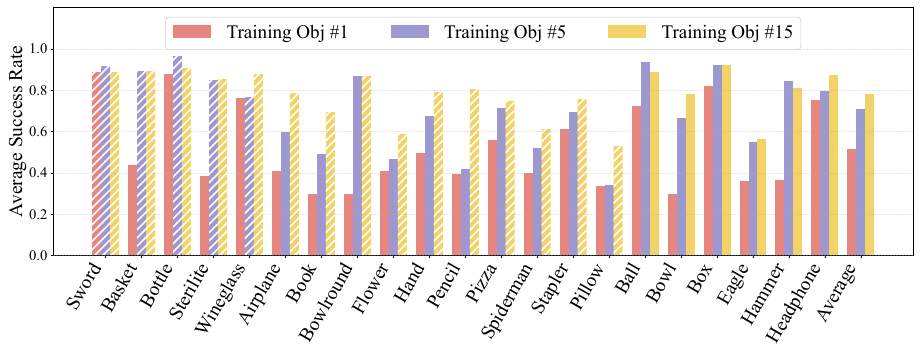}
  \caption{Evaluation of \ours's object generalization performance. The dashed bars denote objects present in the training set, while the undashed bars represent objects unseen during training.}
  \label{fig:obj-generalization}
\end{figure*}

\clearpage
\endgroup

\section{Detailed Sub-Rewards}
\label{sec:detailed-reward}
The unified imitation reward combines complementary components multiplicatively:
\begin{equation}
\begin{aligned}
r_{t} = r_{t}^{hand} r_{t}^{wrist} r_{t}^{object} r_{t}^{interact} r_{t}^{contact},
\end{aligned}
\label{eq:imitation-reward-ap}
\end{equation}

\paragraph{Hand Tracking Reward:}
\begin{equation}
\begin{aligned}
    r_{t}^{hand} = r_{t}^{p} r_{t}^{r},
\end{aligned}
\end{equation}
where $r_{t}^{p}$ denotes the finger joint position reward and $r_{t}^{r}$ represents the finger joint rotation reward.

\paragraph{Wrist Tracking Reward:}
\begin{equation}
\begin{aligned}
    r_{t}^{wrist} = r_{t}^{wp} r_{t}^{wr},
\end{aligned}
\end{equation}
where $r_{t}^{wp}$ denotes the wrist joint position reward and $r_{t}^{wr}$ represents the wrist joint rotation reward.

\paragraph{Object Reward:}
\begin{equation}
\begin{aligned}
    r_{t}^{object} = r_{t}^{op} r_{t}^{or},
\end{aligned}
\end{equation}
where $r_{t}^{op}$ and $r_{t}^{or}$ denote the global object position and rotation rewards, respectively.

\paragraph{Interaction Reward:}
$r_{t}^{interact}$ measures relative hand-object motion by comparing vectors from hand keypoints to object keypoints in the reference data with the corresponding vectors in the simulated environment. This consistency is essential for interaction stability.

\paragraph{Contact Reward:}
$r_{t}^{contact}$ is the contact reward guided by frame-wise binary labels of hand-object contact from reference data, which is crucial for reliable grasping and releasing operations. 

\paragraph{Formulation of Sub-Rewards:}
Each subterm is defined as:
\begin{equation}
    r_{t}^{\alpha} = \exp(-\lambda^{\alpha} e^\alpha_{t+1}),
\end{equation}
where $\alpha \in \{p,r,wp,wr,op,or,interact,contact\}$, and $e_{t+1}^{\alpha}$ and $\lambda^{\alpha}$ denote the tracking error and reward weight, respectively. All reward weights are fixed during training, with $\lambda^{interact}=20$. The values are listed in Table~\ref{tab:params-augmentation}.

%

\section{Domain Randomization Curriculum} 
We apply several domain randomization techniques~\citep{tobin2017domain,peng2018sim,yu2025skillmimic} to enhance policy robustness. During training, we independently randomize three factors: (1) random perturbation forces applied to the object when Move data are sampled; (2) object physical properties, including geometry scale, density, restitution, and friction coefficients; and (3) perturbations to the initial hand-object state. We use a two-stage curriculum with fixed randomization settings within each stage, switching from weaker to stronger randomization at the stage boundary.

\section{Domain Randomization Settings}
\label{sec:dr}

We employ domain randomization along three main axes: object scale, initial state, and a curriculum on observation/action noise.

\paragraph{Object scale randomization.}
During training, the object size is randomly perturbed to expose the policy to a range of geometries. At each environment reset, with probability $0.2$ we uniformly sample a scale factor
\begin{equation}
    s \sim \mathcal{U}(0.75, 1.5),
\end{equation}
and uniformly scale the object by $s$. Otherwise, the object keeps its nominal size. This yields objects between $75\%$ and $150\%$ of the original scale across training episodes, resulting in improved generalization to geometric variations.

\paragraph{Random initial state.}
Beyond sampling a random reference frame from the motion dataset, we additionally inject controlled perturbations into the initial hand-object configuration to further enlarge the distribution of starting states. After selecting a trajectory and a time index, we perturb both the object pose and the hand configuration using zero-mean bounded noise in a phase-dependent manner. For the object, let $\mathbf{p}^o$ and $R^o$ denote the sampled position and orientation. We add a stochastic positional offset
\begin{equation}
    \mathbf{p}^o \leftarrow \mathbf{p}^o + \boldsymbol{\epsilon}_p,
\end{equation}
where the magnitude of $\boldsymbol{\epsilon}_p$ is larger for non-contact frames, encouraging wider spatial variability, and smaller for contact frames to preserve grasp stability. Rotational perturbations follow the same principle: non-contact frames receive full 3D angular noise up to a bounded angle, whereas contact frames use restricted in-plane perturbations to avoid breaking established grasps. For the hand, let $\mathbf{q}$ and $\dot{\mathbf{q}}$ denote joint positions and velocities. Finger-joint angles are perturbed as
\begin{equation}
    \mathbf{q} \leftarrow \mathbf{q} + \boldsymbol{\epsilon}_q,
\end{equation}
where $\boldsymbol{\epsilon}_q$ is sampled within a bounded angular interval to encourage variability in finger postures, and joint velocities receive mild perturbations to model natural initialization variability. For tasks requiring substantial wrist reconfiguration, we also apply small perturbations to the wrist's 6-DoF parameters, providing diverse but physically coherent wrist poses. All perturbations are applied stochastically and independently per environment, ensuring a broad yet physically valid distribution of initial states. Combined with random reference state sampling, this strategy enhances robustness to variations in contact configuration, object placement, and hand posture at the start of each episode.

\paragraph{Randomization curriculum.}
We use a two-stage curriculum with fixed settings within each stage. In the first stage, object-scale randomization is disabled, and initial-state perturbations use a low probability and small magnitude. In the second stage, object-scale randomization is enabled, and both the probability and magnitude of initial-state perturbations switch to higher fixed values.

\section{Adaptive Sampling}
\label{sec:reweight}
To enhance performance on challenging examples, we dynamically adjust sampling probabilities according to sample difficulty~\citep{yu2025skillmimic}. Specifically, for a trajectory $\mathcal{\tau}_i = \{\hat{\boldsymbol{s}}_1, ..., \hat{\boldsymbol{s}}_T\}$, we calculate its mean reward $\bar{r}_i = \frac{1}{T}\sum_{t=1}^{T} r_t$ as a metric of learning difficulty, where $r_t$ is defined in Eq.~\ref{eq:imitation-reward-ap}. Based on this difficulty assessment, we assign a sampling probability $p_i$ for trajectory $\mathcal{\tau}_i$ using:

\begin{equation}
\begin{aligned}
p_i = \frac{e^{-\lambda_s \bar{r}_i}}{\sum_{j=0}^{N-1} e^{-\lambda_s \bar{r}_j}}
\end{aligned}
\end{equation}

where $N$ denotes the dataset size, and $\lambda_s \in [0,\infty)$ controls the balance between uniform sampling ($\lambda_s=0$) and difficulty-prioritized sampling ($\lambda_s>0$).

\section{Implementation Details of Baseline Methods}
\label{sec:method-details}
Because public implementations of SkillMimic~\citep{wang2024skillmimic} and DexGen~\citep{yin2025dexteritygen} do not directly support tracking with the MANO hand model, we construct comparable variants by modifying the reward, input representation, and output format of \ours following the design choices of these methods, as summarized in Tab.~\ref{tab:method-details}. These SkillMimic-style and DexGen-var variants use the same HOG data and training configurations as the specialist \ours policies in Table~\ref{tab:main}.

\begin{table*}[htbp]
  \caption{Implementation details of the compared tracking baselines.}
  \centering
  \vspace{0.2cm}
  \renewcommand{\arraystretch}{1.15}
  \setlength{\tabcolsep}{4pt}
  \resizebox{\linewidth}{!}{
    \begin{tabular}{@{}lcccc@{}}
      \toprule
      \textbf{Feature} & \textbf{\ours} & \textbf{SkillMimic} & \textbf{DexGen-var} & \textbf{DexTrack-style} \\
      \midrule
      \textbf{Observation} & \ours default & No object keypoints & No object information & \ours default \\
      \textbf{Reward} & \ours default & \ours default & \shortstack{Object pose \&\\finger DoF only} & \ours default \\
      \textbf{Action} & \shortstack{Direct finger \&\\residual wrist} & Direct action & Direct action & \shortstack{Residual actions\\for all joints} \\
      \textbf{Data augmentation} & Yes & No & No & No \\
      \textbf{Reweighting} & Yes & No & No & No \\
      \textbf{Homotopy optimization} & No & No & No & 3 rounds \\
      \bottomrule
    \end{tabular}
  }
  \label{tab:method-details}
\end{table*}


\paragraph{DexTrack-style baseline.}
For the unified benchmark, we adapt the action representation and homotopy optimization of DexTrack~\citep{liu2025dextrack} within our codebase. The baseline uses the same observations and rewards as \ours, with residual actions for all joints and the same initial HOG trajectories. Data augmentation and reweighting are disabled. Training proceeds through three stages of 20,000 epochs each. Whenever a reference trajectory is successfully tracked, its simulated rollout replaces the corresponding HOG reference for the next stage. We train separate policies for each skill and object, as for the \ours teachers. This adaptation is labeled DexTrack-style rather than an evaluation of the original implementation. Its total training budget is 60,000 epochs, compared with 30,000 epochs for each \ours teacher; the data-controlled comparison is not compute-matched.

\section{HOG for Other Meta-Skills}
\label{app:other-meta-skills}

\paragraph{Free Move.}
Free Move is an auxiliary non-contact skill for teaching the hand to navigate freely in space. It is not included in the seven interaction-skill evaluations in the main seven-skill benchmark, which focus on hand-object interaction. To generate Free Move data, we randomly select two poses in the operational space as the wrist's initial and final poses. We then randomly generate two sets of finger configurations to form complete initial and final hand states. The trajectory is obtained through interpolation between these states.

\paragraph{Catch.}

Unlike Grasp, the Catch skill involves catching moving objects. To synthesize these demonstrations, we first randomly select a grasp configuration $\boldsymbol{g}$ from $\mathcal{G}$ and generate a parabolic trajectory for the object $[\mathbf{T}^o(t), \mathbf{R}^o(t)] = \text{Parabola}(t), t \in [0,T]$. We then select the last frame as the grasp moment and transform $\boldsymbol{g}$ to align with the object pose $[\mathbf{T}^o(T), \mathbf{R}^o(T)]$, yielding $\boldsymbol{g}^{end}$. The initial hand position and orientation are set to be identical to those in $\boldsymbol{g}^{end}$, while the initial joint angles are set to zero. The hand trajectory is then generated via interpolation between the initial and end hand states. The complete Catch demonstration consists of both the generated hand trajectory and the previously computed parabolic object trajectory. Demonstrations with significant hand-object collisions are filtered out.

\paragraph{Throw.}
Throw demonstrations, which involve throwing objects with a specified velocity, are generated by time-reversing Catch trajectories. The parabolic motion of a thrown object remains physically reasonable due to the symmetric structure of the trajectory.

\paragraph{General Rotate.}
\label{app:general-rotate}
In addition to the simplified Rotate data generation method described in the main text, we also propose a more general approach for generating rotation data. This method produces more challenging data that require longer training times. Specifically, we first select a grasp configuration $\boldsymbol{g}^1$ from $\mathcal{G}$ as the initial state and remove it from $\mathcal{G}$. We then iteratively find the closest grasp configuration in the remaining set: for each $\boldsymbol{g}^i$, we search for its nearest grasp configuration $\boldsymbol{g}^{i+1}$ and remove $\boldsymbol{g}^{i+1}$ from $\mathcal{G}$, continuing this process for $k$ steps. We transform all grasp configurations to align their wrist poses with that of $\boldsymbol{g}^1$. To provide sufficient buffer time to learn such large transitions, we replicate each $\boldsymbol{g}^i$ for $N$ frames.

\paragraph{Regrasp.}
We define Regrasp as a skill that differs from Rotate in that it maintains a fixed object pose while transitioning between different hand configurations. The generation methodology mirrors that of Rotate, with the key distinction being that the grasp configurations are aligned to the same object pose rather than the same wrist pose.

\paragraph{Retargeting GRAB for Evaluation.}
\label{sec:retarget-grab}
We validated the system's ability to track real-world HOI trajectories using sequences from the GRAB dataset~\citep{GRAB2020}. 
These were segmented into individual grasp-move-place episodes. An initial hand trajectory was obtained by retargeting based on the ManipTrans~\citep{li2025maniptrans} method, providing preliminary wrist and joint configurations.
We further employed HOG to synthesize an improved grasp pose by fixing the object and wrist rotations from the initial grasp frame and optimizing the hand configuration toward higher grasp quality and physical plausibility while preserving kinematic similarity. The original finger and object data in each trajectory were replaced with this refined pose, yielding a corrected GRAB dataset that serves as a high-quality benchmark for evaluation.

\begin{figure}[t]
  \centering  \includegraphics[width=\linewidth]{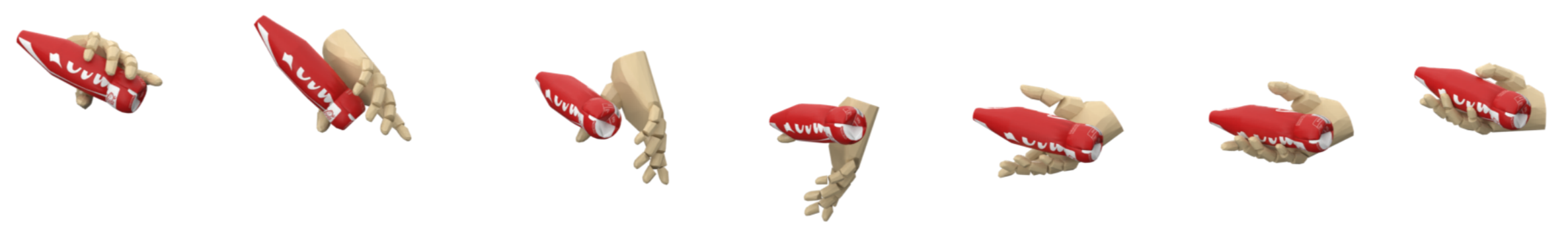}
  \caption{Tracking a Regrasp trajectory.} 
  \label{fig:regrasp}
\end{figure}

\section{LLM-based Trajectory Planning}
\label{sec:interpolation}

We employ an LLM to generate semantically annotated keyframes and subsequently apply keyframe-guided interpolation in both position and orientation. This pipeline ensures that the generated trajectories maintain both semantic coherence and dynamic feasibility, making them directly applicable to our tracking policy for practical manipulation tasks.
\paragraph{LLM Prompt and Keypoint Specification.}
Given a high-level manipulation command (e.g., ``set a lying bottle upright''), we construct a structured prompt that includes the task description, the initial wrist and object poses in a fixed robot base frame, the coordinate system definition, rotation conventions in XYZ Euler angles and quaternions, and smoothness constraints on positional and rotational changes.
The prompt also provides a full desktop configuration listing all objects together with their feasible candidate grasp configurations, each associated with a valid wrist pose. The LLM is explicitly required to \emph{select one} candidate grasp and align the generated grasp keypoint to that wrist pose.

The LLM outputs a JSON object containing keypoints and metadata. Each keypoint specifies a unique index, wrist position $\mathbf{p}_i$, orientation $q_i$, and an action label. Keypoints with \textit{grasp} and \textit{release} actions must be designated, with their indices explicitly provided in top-level fields. These frames define interpolation segments, while start/end frames are derived from sequence boundaries. Other keypoints serve as internal anchors.

\paragraph{Keyframe-Based Piecewise Bezier Interpolation.}
The four special frame types naturally partition the LLM keypoints into three semantic segments: start $\rightarrow$ grasp, grasp $\rightarrow$ release, and release $\rightarrow$ end. Within each segment, all keypoints (including intermediate approach/transport points) are used as interpolation anchors, while different segments are interpolated independently. For each segment, we construct a cubic Bezier curve for the wrist position using the segment endpoints as Bezier endpoints and generating two inner control points from finite differences of neighboring anchors, so that the curve tangents align with the discrete motion directions proposed by the LLM. We then sample a fixed number of trajectory points per LLM keypoint and concatenate all segments into a dense position trajectory, removing duplicated boundary samples. This piecewise scheme ensures smooth motion inside each phase while strictly preserving the grasp and release boundaries (e.g., if the LLM predicts indices 3 and 8 as grasp and release, we only interpolate within [1,3], [3,8], and [8,9]).

\paragraph{Orientation Interpolation.}
Wrist orientations are interpolated independently of positions. We first convert all keyframe orientations to quaternions, enforce sign continuity along the sequence to avoid double-cover flipping, and then apply spherical linear interpolation (SLERP) within each segment using the same temporal parameterization as for the position Bezier curves. Segments with negligible rotation are treated as static segments, for which the orientation is kept fixed throughout the segment.

The trajectories generated by both the LLM and Text2HOI~\citep{cha2024text2hoi} are subsequently processed by HOG, which replaces the original finger degrees of freedom with synthesized grasp poses.

\section{Policy Distillation}
\label{app:distillation}

\begin{table*}[t]
  \caption{Additional four-skill distillation experiment with its own student, distinct from the seven-skill and 15-object students.}
  \label{tab:multi_task_distill}
  \centering
  \renewcommand\arraystretch{1.2}
  \resizebox{\linewidth}{!}{%
    \begin{tabular}{llcccc@{\hspace{1.5em}}cccc}
      \toprule
      \multirow{2}{*}{Skill} & \multirow{2}{*}{Method}
        & \multicolumn{4}{c}{Train} & \multicolumn{4}{c}{Test} \\
      \cmidrule(lr){3-6} \cmidrule(lr){7-10}
        & & SR (\%) $\uparrow$ & $E_{op}$ (cm) $\downarrow$ & $E_{or}$ (${}^\circ$) $\downarrow$ & $E_h$ (cm) $\downarrow$
          & SR (\%) $\uparrow$ & $E_{op}$ (cm) $\downarrow$ & $E_{or}$ (${}^\circ$) $\downarrow$ & $E_h$ (cm) $\downarrow$ \\
      \midrule
      \multirow{3}{*}{Grasp} & Teacher & 98.19 & 1.03 & 6.20 & 3.74 & 73.89 & 3.21 & 22.93 & 5.66 \\
       & Naive Multi-Task & 66.01 & 4.57 & 16.61 & 7.82 & 68.38 & 5.32 & 17.39 & 7.59 \\
       & Distilled Student & 97.88 & 0.86 & 4.85 & 3.06 & 79.72 & 2.50 & 20.76 & 4.81 \\
      \midrule
      \multirow{3}{*}{Move} & Teacher & 99.29 & 2.31 & 10.57 & 4.12 & 99.14 & 2.50 & 10.85 & 5.16 \\
       & Naive Multi-Task & 81.40 & 4.57 & 16.61 & 7.82 & 89.45 & 3.17 & 11.76 & 5.56 \\
       & Distilled Student & 98.25 & 1.97 & 10.45 & 4.43 & 97.71 & 2.27 & 10.68 & 3.95 \\
      \midrule
      \multirow{3}{*}{Place} & Teacher & 98.41 & 1.04 & 6.02 & 3.04 & 98.13 & 1.09 & 8.42 & 2.85 \\
       & Naive Multi-Task & 82.63 & 3.49 & 29.02 & 5.77 & 84.88 & 3.47 & 27.83 & 5.84 \\
       & Distilled Student & 97.29 & 1.15 & 7.39 & 3.03 & 97.32 & 1.04 & 8.67 & 2.66 \\
      \midrule
      \multirow{3}{*}{Rotate} & Teacher & 97.30 & 0.98 & 7.75 & 1.71 & 56.23 & 2.38 & 44.96 & 2.57 \\
       & Naive Multi-Task & 2.42 & 2.27 & 27.67 & 4.75 & 0.39 & 3.00 & 27.97 & 5.06 \\
       & Distilled Student & 83.76 & 1.49 & 11.23 & 2.28 & 45.44 & 2.42 & 56.61 & 2.71 \\
      \bottomrule
    \end{tabular}
  }
\end{table*}

\paragraph{Distinct policy settings.}
The main seven-skill benchmark uses one student distilled across Grasp, Move, Place, Throw, Catch, Rotate, and Regrasp on four objects. The object-generalization experiment uses a separately trained student distilled from object-specific teachers on Grasp, Move, and Place across 15 objects; this is the student used for the 67.27\% frame-wise SR on unseen Grasp--Move--Place compositions. The four-skill experiment in Table~\ref{tab:multi_task_distill} is an additional distillation ablation, with its own student. These settings do not share a single student checkpoint.

To address skill interference in multi-task and multi-object learning, we implement a progressive knowledge distillation framework that transfers expertise from specialized teacher policies to a unified student policy. Inspired by previous work~\citep{luo2023perpetual,xu2025intermimic}, we employ a four-stage training strategy: 
\textbf{Stage I} (0--500 epochs) uses the DAgger algorithm for behavioral cloning, where actions are sampled entirely from teacher policies and the student policy is updated using an expert supervision loss (ELoss). 
\textbf{Stage II} (500--5000 epochs) introduces hybrid decision-making through expert-mode masking, with the teacher action sampling probability linearly decaying from 100\% to 0\% to enable a smooth transition to autonomous policy execution. 
\textbf{Stage III} (5000--7000 epochs) focuses on value function optimization, progressively incorporating the policy gradient loss when explained variance (EV) consistently exceeds 0.6 for three consecutive evaluation windows. 
\textbf{Stage IV} (7000+ epochs) combines policy gradient, value function, boundary constraint, and residual expert supervision losses, with dynamic weights determined by training progress and critic performance.

To evaluate the effectiveness of policy distillation, we compare the performance of teacher policies, the distilled student policy, and a naively trained multi-task policy in an additional four-skill tracking experiment covering Grasp, Move, Place, and Rotate. As shown in Tab.~\ref{tab:multi_task_distill}, the distilled student policy mitigates skill interference commonly observed in multi-task learning, raising Rotate test SR from 0.39\% to 45.44\% (a gain of 45.05 percentage points). Moreover, the distillation process reduces the required training cycles by approximately 45\% while achieving higher final performance compared to naive multi-task training.

To evaluate how distillation improves generalization across objects, we conduct experiments on the Grasp, Move, and Place skill data under several training settings:
a student policy distilled without RL from 15 objects; a student policy using both distillation and RL with 15 objects; a Distill\&RL policy trained on 5 objects; and a policy trained on a single object (Sword).
As shown in Tab.~\ref{tab:merged_policy_table}, the Distill\&RL approach yields clear gains over distillation alone. Moreover, performance improves consistently as the number of training objects increases, confirming the benefit of multi-object training for generalization.

\begin{table*}[t]
  \centering
  \begin{minipage}[t]{0.74\textwidth}
    \vspace{0pt}
    \caption{Multi-object distillation on Grasp, Move, and Place. These students are trained separately from the seven-skill student.}
    \renewcommand{\arraystretch}{1.35}
    \centering
    \resizebox{\linewidth}{!}{%
      \begin{tabular}{l|c|c|c|c}
        \toprule
        \multirow{3}{*}{Object}
          & \multicolumn{1}{c|}{\textbf{Distillation Only}}
          & \multicolumn{1}{c|}{\textbf{Distillation + RL}}
          & \multicolumn{1}{c|}{\textbf{Distillation + RL}}
          & \multicolumn{1}{c}{\textbf{Sword Teacher}} \\
          & \multicolumn{1}{c|}{\textbf{(15 Objects)}}
          & \multicolumn{1}{c|}{\textbf{(15 Objects)}}
          & \multicolumn{1}{c|}{\textbf{(5 Objects)}}
          & \multicolumn{1}{c}{\textbf{(1 Object)}} \\
        \cmidrule(lr){2-2}
        \cmidrule(lr){3-3}
        \cmidrule(lr){4-4}
        \cmidrule(lr){5-5}
          & Grasp/Move/Place SR $\uparrow$ 
          & Grasp/Move/Place SR $\uparrow$
          & Grasp/Move/Place SR $\uparrow$
          & Grasp/Move/Place SR $\uparrow$ \\
        \midrule
        Airplane & 39\%\enspace/\enspace 60\%\enspace/\enspace 50\% & \textbf{62\%}\enspace/\enspace \textbf{84\%}\enspace/\enspace \textbf{89\%} & 59\%\enspace/\enspace 66\%\enspace/\enspace 54\% & 44\%\enspace/\enspace 46\%\enspace/\enspace 32\% \\
        Ball & 60\%\enspace/\enspace 91\%\enspace/\enspace 90\% & 79\%\enspace/\enspace 88\%\enspace/\enspace \textbf{100\%} & \textbf{96\%}\enspace/\enspace 85\%\enspace/\enspace \textbf{100\%} & 46\%\enspace/\enspace \textbf{99\%}\enspace/\enspace 72\% \\
        Basket & 56\%\enspace/\enspace 91\%\enspace/\enspace \textbf{99\%} & 76\%\enspace/\enspace \textbf{98\%}\enspace/\enspace 93\% & \textbf{87\%}\enspace/\enspace 82\%\enspace/\enspace 98\% & 38\%\enspace/\enspace 57\%\enspace/\enspace 36\% \\
        Book & 14\%\enspace/\enspace 59\%\enspace/\enspace 41\% & \textbf{51\%}\enspace/\enspace \textbf{83\%}\enspace/\enspace \textbf{74\%} & 33\%\enspace/\enspace 53\%\enspace/\enspace 61\% & 15\%\enspace/\enspace 45\%\enspace/\enspace 30\% \\
        Bottle & 65\%\enspace/\enspace 94\%\enspace/\enspace 96\% & 78\%\enspace/\enspace 95\%\enspace/\enspace 98\% & \textbf{89\%}\enspace/\enspace \textbf{100\%}\enspace/\enspace \textbf{100\%} & 75\%\enspace/\enspace 95\%\enspace/\enspace 93\% \\
        Bowl & 38\%\enspace/\enspace 78\%\enspace/\enspace 74\% & \textbf{76\%}\enspace/\enspace \textbf{79\%}\enspace/\enspace \textbf{80\%} & 62\%\enspace/\enspace 76\%\enspace/\enspace 62\% & 22\%\enspace/\enspace 56\%\enspace/\enspace 11\% \\
        Bowlround & 64\%\enspace/\enspace 88\%\enspace/\enspace 84\% & \textbf{86\%}\enspace/\enspace 89\%\enspace/\enspace \textbf{85\%} & 84\%\enspace/\enspace \textbf{91\%}\enspace/\enspace \textbf{85\%} & 38\%\enspace/\enspace 43\%\enspace/\enspace 8\% \\
        Box & 55\%\enspace/\enspace 97\%\enspace/\enspace 95\% & \textbf{78\%}\enspace/\enspace 98\%\enspace/\enspace \textbf{100\%} & 77\%\enspace/\enspace \textbf{100\%}\enspace/\enspace \textbf{100\%} & 57\%\enspace/\enspace \textbf{100\%}\enspace/\enspace 89\% \\
        Eagle & 28\%\enspace/\enspace \textbf{67\%}\enspace/\enspace 42\% & 45\%\enspace/\enspace 51\%\enspace/\enspace \textbf{73\%} & \textbf{46\%}\enspace/\enspace \textbf{67\%}\enspace/\enspace 52\% & 28\%\enspace/\enspace 53\%\enspace/\enspace 27\% \\
        Flower & 22\%\enspace/\enspace 62\%\enspace/\enspace 53\% & \textbf{44\%}\enspace/\enspace \textbf{65\%}\enspace/\enspace \textbf{66\%} & 35\%\enspace/\enspace 56\%\enspace/\enspace 49\% & 26\%\enspace/\enspace 58\%\enspace/\enspace 38\% \\
        Hammer & 20\%\enspace/\enspace 88\%\enspace/\enspace 20\% & 67\%\enspace/\enspace 85\%\enspace/\enspace 91\% & \textbf{70\%}\enspace/\enspace \textbf{90\%}\enspace/\enspace \textbf{94\%} & 17\%\enspace/\enspace 87\%\enspace/\enspace 6\% \\
        Hand & 37\%\enspace/\enspace 81\%\enspace/\enspace 73\% & \textbf{66\%}\enspace/\enspace \textbf{83\%}\enspace/\enspace \textbf{87\%} & 62\%\enspace/\enspace 74\%\enspace/\enspace 67\% & 34\%\enspace/\enspace 64\%\enspace/\enspace 51\% \\
        Headphones & 53\%\enspace/\enspace 93\%\enspace/\enspace 89\% & \textbf{71\%}\enspace/\enspace \textbf{97\%}\enspace/\enspace \textbf{94\%} & 68\%\enspace/\enspace 87\%\enspace/\enspace 84\% & 64\%\enspace/\enspace 86\%\enspace/\enspace 75\% \\
        Pencil & 37\%\enspace/\enspace 67\%\enspace/\enspace 79\% & \textbf{67\%}\enspace/\enspace \textbf{94\%}\enspace/\enspace \textbf{80\%} & 37\%\enspace/\enspace 58\%\enspace/\enspace 31\% & 23\%\enspace/\enspace 50\%\enspace/\enspace 45\% \\
        Pillow & 11\%\enspace/\enspace 52\%\enspace/\enspace 44\% & \textbf{30\%}\enspace/\enspace \textbf{61\%}\enspace/\enspace \textbf{67\%} & 14\%\enspace/\enspace 45\%\enspace/\enspace 43\% & 9\%\enspace/\enspace 42\%\enspace/\enspace 50\% \\
        Pizza & \textbf{48\%}\enspace/\enspace 70\%\enspace/\enspace 75\% & 44\%\enspace/\enspace \textbf{100\%}\enspace/\enspace 79\% & 46\%\enspace/\enspace 74\%\enspace/\enspace \textbf{94\%} & \textbf{48\%}\enspace/\enspace 54\%\enspace/\enspace 65\% \\
        Spidermanfigure & 31\%\enspace/\enspace 69\%\enspace/\enspace 45\% & \textbf{59\%}\enspace/\enspace 53\%\enspace/\enspace \textbf{71\%} & 40\%\enspace/\enspace \textbf{70\%}\enspace/\enspace 46\% & 29\%\enspace/\enspace 61\%\enspace/\enspace 30\% \\
        Stapler & 37\%\enspace/\enspace 78\%\enspace/\enspace \textbf{80\%} & \textbf{62\%}\enspace/\enspace \textbf{86\%}\enspace/\enspace 78\% & 60\%\enspace/\enspace 72\%\enspace/\enspace 76\% & 41\%\enspace/\enspace 72\%\enspace/\enspace 71\% \\
        Sterilite & 46\%\enspace/\enspace 91\%\enspace/\enspace 84\% & 73\%\enspace/\enspace 88\%\enspace/\enspace \textbf{95\%} & \textbf{79\%}\enspace/\enspace \textbf{94\%}\enspace/\enspace 81\% & 35\%\enspace/\enspace 65\%\enspace/\enspace 15\% \\
        Sword & 71\%\enspace/\enspace 96\%\enspace/\enspace 94\% & 68\%\enspace/\enspace \textbf{100\%}\enspace/\enspace 98\% & 76\%\enspace/\enspace 98\%\enspace/\enspace \textbf{100\%} & \textbf{79\%}\enspace/\enspace 99\%\enspace/\enspace 88\% \\
        Wineglass & 65\%\enspace/\enspace 74\%\enspace/\enspace 82\% & \textbf{90\%}\enspace/\enspace \textbf{89\%}\enspace/\enspace \textbf{84\%} & 80\%\enspace/\enspace 71\%\enspace/\enspace 78\% & 70\%\enspace/\enspace 80\%\enspace/\enspace 79\% \\
        \midrule
        Average & 43\%\enspace/\enspace 78\%\enspace/\enspace 71\% & \textbf{65\%}\enspace/\enspace \textbf{84\%}\enspace/\enspace \textbf{85\%} & 62\%\enspace/\enspace 77\%\enspace/\enspace 74\% & 40\%\enspace/\enspace 67\%\enspace/\enspace 48\% \\
        \bottomrule
      \end{tabular}
    }
    \label{tab:merged_policy_table}
  \end{minipage}\hfill
  \begin{minipage}[t]{0.26\textwidth}
    \vspace{0pt}
    \centering
    \includegraphics[width=\linewidth]{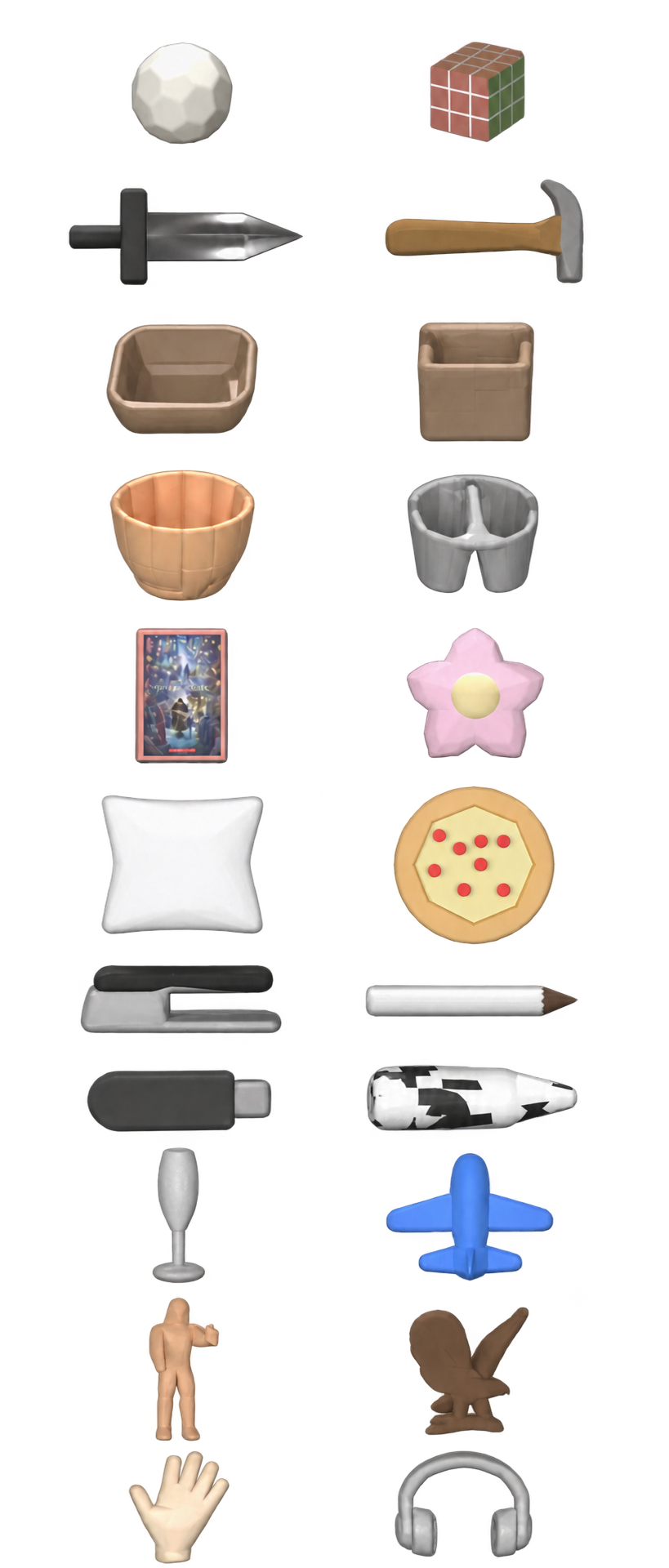}
    \label{fig:obj-generalization-qualitative}
  \end{minipage}
\end{table*}

\section{Hyperparameters}
\label{sec:hyperparameters}
We summarize the simulation parameters, training hyperparameters, reward coefficients, and data augmentation settings in Tab.~\ref{tab:params-augmentation}.

\begin{table*}[h]
  \caption{Hyperparameters for policy training, imitation rewards, data augmentation, and adaptive sampling. For entries with two values, the first applies to Regrasp and the second to all other skills. Joint and object disturbances are specified by their amplitudes.}
  \label{tab:params-augmentation}
  \centering
  \scriptsize
  \begin{minipage}[t]{0.32\linewidth}
    \renewcommand{\arraystretch}{1.1}
    \centering
    \textbf{Training}\par\smallskip
    \begin{tabular}{@{}p{0.66\linewidth}@{\hspace{3pt}}c@{}}
      \toprule
      \textbf{Parameter} & \textbf{Value} \\
      \midrule
      Action variance $\Sigma_{\boldsymbol{\pi}}$ & 0.055 \\
      Samples per update & 65,536 \\
      Policy/value minibatch size & 16,384 \\
      Discount $\gamma$ & 0.99 \\
      Adam learning rate & $2\times10^{-5}$ \\
      GAE($\lambda$) & 0.95 \\
      TD($\lambda$) & 0.95 \\
      PPO clip threshold & 0.2 \\
      Episode length & 60 \\
      \bottomrule
    \end{tabular}
  \end{minipage}\hfill
  \begin{minipage}[t]{0.29\linewidth}
   \renewcommand{\arraystretch}{1.25}
    \centering
    \textbf{Reward Coefficients}\par\smallskip
    \begin{tabular}{@{}p{0.64\linewidth}@{\hspace{3pt}}c@{}}
      \toprule
      \textbf{Term} & \textbf{Weight} \\
      \midrule
      Position $\lambda^{p}$ & $200 / 20$ \\
      Rotation $\lambda^{r}$ & $200 / 20$ \\
      Object position $\lambda^{op}$ & 50 \\
      Object rotation $\lambda^{or}$ & 50 \\
      Interaction $\lambda^{interact}$ & 20 \\
      Contact $\lambda^{contact}$ & 5 \\
      Wrist position $\lambda^{wp}$ & 20 \\
      Wrist rotation $\lambda^{wr}$ & 20 \\
      \bottomrule
    \end{tabular}
  \end{minipage}\hfill
  \begin{minipage}[t]{0.35\linewidth}
    \renewcommand{\arraystretch}{1.6}
    \centering
    \textbf{Augmentation \& Sampling}\par\smallskip
    \begin{tabular}{@{}p{0.68\linewidth}@{\hspace{3pt}}c@{}}
      \toprule
      \textbf{Parameter} & \textbf{Value} \\
      \midrule
      Initial disturbance probability & $0.5 / 0.3$ \\
      Joint position disturbance & $\pi/8$ \\
      Joint velocity disturbance & 0.1 \\
      Object velocity disturbance & 0.02 \\
      Object rotation disturbance & 0.02 \\
      Reweighting coefficient $\lambda^{s}$ & 10 \\
      \bottomrule
    \end{tabular}
  \end{minipage}
\end{table*}
\section{Following Text Instructions}

Our framework supports manipulation from natural language instructions through multiple complementary approaches. The typical pipeline uses a Vision-Language Model (VLM) for high-level semantic planning. The VLM takes the following inputs: the object's initial 6D pose, a set of feasible wrist poses corresponding to stable grasp configurations, the initial wrist pose, the target object position, an image of the current environment, and a natural-language task description. It outputs a structured plan containing (1) sparse 6D waypoints for the wrist at key states (start, grasp, release, end), (2) a selected grasp pose, and (3) binary flags indicating grasp and release events.
We also explore an alternative that uses Text2HOI \citep{cha2024text2hoi} to generate wrist trajectories directly from language. In both cases, we apply trajectory smoothing and Bezier interpolation between keyframes to ensure smooth motions.
The resulting wrist path is combined with the chosen grasp pose using the procedure in Sec.~\ref{sec: traj synth} of the main text to form a complete hand-object demonstration. This is then tracked by our distilled \ours student, demonstrating end-to-end execution of language-guided manipulation. Fig.~\ref{fig: teaser} in the main text shows representative results.

\section{Unified Seven-Skill Benchmark}
\label{app:unified-benchmark}
\suppressfloats[t]

\paragraph{Protocol and policy scope.}
The unified benchmark uses the MANO hand and four objects: a box, a bottle, a sword, and a hammer. For each skill and object, DexTrack-style and \ours teacher results come from separately trained specialist policies. The seven-skill \ours student is one policy shared across all seven skills and four objects. Each object--skill setting is evaluated using five random seeds with 1,000 trials per seed. Table~\ref{tab:main} reports the mean and standard deviation from these evaluations, with results averaged over the four objects. The reported variability concerns evaluation seeds, not independent training runs.

\paragraph{Reward trade-offs.}
Contact supervision is especially important for Grasp, Throw, and Catch, while interaction supervision has a large effect on Rotate and Regrasp. Removing either term can improve Place SR, so these ablations support skill-dependent benefits rather than uniform gains. These are teacher-policy ablations, not student ablations.

\section{Evaluation of HOI Data Quality and Coverage}
\label{app:data-evaluation}

\subsection{Evaluation Data}
We evaluate data quality and coverage on two object categories: bowls and cylinders. Each method contributes 20 hand-object interaction trajectories. EgoGrasp trajectories are reconstructed from captured footage; LatentHOI~\citep{li2025latenthoi} provides 10 generated trajectories per object category; GRAB trajectories are selected from the original dataset; and DexMimicGen and HOG trajectories are randomly sampled from their respective outputs.

\subsection{Data Quality Metrics}
\paragraph{Penetration.}
Penetration measures the average depth of intersection between the hand and object meshes, reported in millimeters. Lower values indicate less interpenetration.

\paragraph{Jittering.}
Jittering measures high-frequency pose oscillations, reported in millimeters. Lower values indicate smoother motion.

\paragraph{Grasp Success Rate (GSR).}
GSR is the percentage of simulated hold tests completed without dropping the object:
\begin{equation}
    \mathrm{GSR} = \frac{N_{\mathrm{successful\ holds}}}{N_{\mathrm{hold\ tests}}} \times 100\%.
\end{equation}

\subsection{Data Coverage Metrics}
\paragraph{Spatial Coverage.}
The State-Space Coverage column in Table~\ref{tab:data_coverage} measures spatial coverage using the volume of the three-dimensional axis-aligned bounding box enclosing all wrist positions across the evaluated trajectories. Let $\mathcal{P}$ denote these wrist positions. The coverage is
\begin{equation}
    C_{\mathrm{spatial}} = \prod_{a\in\{x,y,z\}}
    \left(\max_{\boldsymbol{p}\in\mathcal{P}} p_a - \min_{\boldsymbol{p}\in\mathcal{P}} p_a\right).
\end{equation}
Larger values indicate that the trajectories explore a wider workspace.

\paragraph{Grasp Diversity.}
Grasp diversity is the average $k$-nearest-neighbor distance between grasp descriptors expressed in the object coordinate frame. Each descriptor includes the grasp translation, rotation, and hand joint configuration. For descriptors $\{\boldsymbol{g}_i\}_{i=1}^{N}$, the score is
\begin{equation}
    D_{\mathrm{grasp}} = \frac{1}{Nk}\sum_{i=1}^{N}
    \sum_{j\in\mathcal{N}_k(i)} d(\boldsymbol{g}_i,\boldsymbol{g}_j),
\end{equation}
where $\mathcal{N}_k(i)$ contains the $k$ nearest descriptors other than $\boldsymbol{g}_i$, and $d$ denotes the descriptor distance. Larger scores indicate more diverse grasps.


\section{Limitations and Discussion}

\label{sec:limitations}

While our method presents a promising path to addressing the data bottleneck in dexterous manipulation, several limitations remain. First, although \ours provides a general-purpose low-level control capability for trajectory tracking, and we have demonstrated its integration with high-level planners such as LLMs \citep{achiam2023gpt} and Text2HOI \citep{cha2024text2hoi}, the current high-level planners operate in an open-loop manner. They do not yet support the kind of long-horizon, closed-loop task execution provided by an LLM-based agent. For example, when an object is dropped, the system lacks the capability to autonomously regenerate a recovery trajectory. Achieving such robustness requires both more adaptive tracking policies and more reactive, long-horizon planners.

Second, the unified seven-skill student is evaluated on four objects and loses performance relative to specialist teachers on six of the seven skills. Our 15-object student is a separately trained model covering only Grasp, Move, and Place. Thus, the two experiments do not establish one policy that jointly scales to all seven skills and a large object set. The 67.27\% frame-wise SR applies only to zero-shot Grasp--Move--Place composition on four objects; the more complex long-horizon examples use separately trained composite-task policies. Skills such as Regrasp, while tractable for individual objects (see Fig.~\ref{fig:regrasp}), become increasingly difficult to learn as object diversity grows. Scaling to more complex skills and broader object categories may require large-scale, learning-based refinement of HOG data and hierarchical distillation strategies.

Furthermore, real-world deployment faces the sim-to-real gap, particularly for the highly dynamic motions that our method excels at in simulation. This demands hardware with high control frequency and force-sensing capabilities—features that remain limited in current robotic hands.

Overall, this work demonstrates the feasibility of a general-purpose dexterous controller learned from synthetic data, yet several open challenges remain for further scaling and real-world application.

\clearpage
\section*{NeurIPS Paper Checklist}

\begin{enumerate}

\item {\bf Claims}
    \item[] Question: Do the main claims made in the abstract and introduction accurately reflect the paper's contributions and scope?
    \item[] Answer: \answerYes{} 
    \item[] Justification: The abstract and introduction summarize the contributions on synthetic trajectory generation, tracking, and policy distillation, and these claims are supported by experimental evaluations and ablations.

\item {\bf Limitations}
    \item[] Question: Does the paper discuss the limitations of the work performed by the authors?
    \item[] Answer: \answerYes{} 
    \item[] Justification: The appendix discusses open-loop high-level planning, skill interference in multi-task and multi-object learning, and sim-to-real challenges.

\item {\bf Theory assumptions and proofs}
    \item[] Question: For each theoretical result, does the paper provide the full set of assumptions and a complete (and correct) proof?
    \item[] Answer: \answerNA{} 
    \item[] Justification: This paper is primarily empirical and focuses on robotic control algorithms and data generation; it does not include new theoretical proofs.

    \item {\bf Experimental result reproducibility}
    \item[] Question: Does the paper fully disclose all the information needed to reproduce the main experimental results of the paper to the extent that it affects the main claims and/or conclusions of the paper (regardless of whether the code and data are provided or not)?
    \item[] Answer:  \answerYes{} 
    \item[] Justification: The manuscript and the public \href{https://github.com/wyhuai/Hand-Object-Tracking}{code repository} provide implementation details, synthetic motion data, pretrained teacher checkpoints, environment setup, and instructions for data generation, \ours training, evaluation, and policy distillation. However, the release documentation does not yet provide a complete experiment-by-experiment reproduction guide for all reported baseline comparisons and ablations.

\item {\bf Open access to data and code}
    \item[] Question: Does the paper provide open access to the data and code, with sufficient instructions to faithfully reproduce the main experimental results, as described in supplemental material?
    \item[] Answer: \answerYes{}
    \item[] Justification: Code, synthetic motion data, and pretrained teacher checkpoints are publicly available through \url{https://github.com/wyhuai/Hand-Object-Tracking}, including a link to the full motion dataset. The repository provides installation and data preparation instructions, data generation code, and commands for \ours training, evaluation, and distillation; it documents the supported workflows, while a complete reproduction guide for all baseline comparisons and ablations is not yet provided.

\item {\bf Experimental setting/details}
    \item[] Question: Does the paper specify all the training and test details (e.g., data splits, hyperparameters, how they were chosen, type of optimizer) necessary to understand the results?
    \item[] Answer: \answerYes{} 
    \item[] Justification: The main paper specifies the evaluation protocol, and the appendix provides domain randomization settings, baseline implementation details, policy distillation details, and hyperparameter configurations needed to understand the results.

\item {\bf Experiment statistical significance}
    \item[] Question: Does the paper report error bars suitably and correctly defined or other appropriate information about the statistical significance of the experiments?
    \item[] Answer: \answerYes{} 
    \item[] Justification: The core seven-skill comparison and teacher reward ablations (Table~\ref{tab:main}) report means and standard deviations from five evaluation seeds, with 1,000 trials per object--skill setting and seed. These measure evaluation variability, not variability across independently trained policies. Other experiments do not systematically report uncertainty, and we do not claim formal statistical significance.

\item {\bf Experiments compute resources}
    \item[] Question: For each experiment, does the paper provide sufficient information on the computer resources (type of compute workers, memory, time of execution) needed to reproduce the experiments?
    \item[] Answer: \answerNo{}
    \item[] Justification: We specify the main hardware platform and training schedules, but the paper does not report detailed memory usage, wall-clock time, or total compute estimates for all experiments.
    
\item {\bf Code of ethics}
    \item[] Question: Does the research conducted in the paper conform, in every respect, with the NeurIPS Code of Ethics \url{https://neurips.cc/public/EthicsGuidelines}?
    \item[] Answer: \answerYes{} 
    \item[] Justification: The research was conducted in compliance with the NeurIPS Code of Ethics using simulation, synthetic data, and publicly available benchmark data.

\item {\bf Broader impacts}
    \item[] Question: Does the paper discuss both potential positive societal impacts and negative societal impacts of the work performed?
    \item[] Answer: \answerNo{} 
    \item[] Justification: The paper focuses on technical contributions to simulated dexterous manipulation and does not include a dedicated discussion of both positive and negative societal impacts.
    
\item {\bf Safeguards}
    \item[] Question: Does the paper describe safeguards that have been put in place for responsible release of data or models that have a high risk for misuse (e.g., pre-trained language models, image generators, or scraped datasets)?
    \item[] Answer: \answerNA{} 
    \item[] Justification: This research focuses on physical simulation for robot control and does not pose risks associated with high-stakes generative models or sensitive data.

\item {\bf Licenses for existing assets}
    \item[] Question: Are the creators or original owners of assets (e.g., code, data, models), used in the paper, properly credited and are the license and terms of use explicitly mentioned and properly respected?
    \item[] Answer: \answerNo{} 
    \item[] Justification: We cite the sources of the existing datasets, simulation tools, and baseline methods, and our released code repository includes an Apache-2.0 license. However, the manuscript does not comprehensively list the licenses and terms of use for the third-party assets used in this work.

\item {\bf New assets}
    \item[] Question: Are new assets introduced in the paper well documented and is the documentation provided alongside the assets?
    \item[] Answer: \answerYes{} 
    \item[] Justification: The public repository provides a \href{https://github.com/wyhuai/Hand-Object-Tracking#readme}{README} and a \href{https://github.com/wyhuai/Hand-Object-Tracking/blob/main/DATA_GENERATION.md}{data generation guide} documenting the released code, synthetic motion data, and pretrained teacher checkpoints, including installation, data organization and generation, training, evaluation, and distillation. The code repository includes an Apache-2.0 license, and methodological limitations are discussed in Appendix~\ref{sec:limitations}.

\item {\bf Crowdsourcing and research with human subjects}
    \item[] Question: For crowdsourcing experiments and research with human subjects, does the paper include the full text of instructions given to participants and screenshots, if applicable, as well as details about compensation (if any)? 
    \item[] Answer: \answerNA{} 
    \item[] Justification: This research did not conduct new crowdsourcing experiments or human-subject studies; existing public human-motion data are used only as a benchmark.

\item {\bf Institutional review board (IRB) approvals or equivalent for research with human subjects}
    \item[] Question: Does the paper describe potential risks incurred by study participants, whether such risks were disclosed to the subjects, and whether Institutional Review Board (IRB) approvals (or an equivalent approval/review based on the requirements of your country or institution) were obtained?
    \item[] Answer: \answerNA{} 
    \item[] Justification: No new human-subject study, crowdsourcing study, or sensitive personal data collection was conducted in this research.

\item {\bf Declaration of LLM usage}
    \item[] Question: Does the paper describe the usage of LLMs if it is an important, original, or non-standard component of the core methods in this research? Note that if the LLM is used only for writing, editing, or formatting purposes and does \emph{not} impact the core methodology, scientific rigor, or originality of the research, declaration is not required.
    \item[] Answer: \answerYes{} 
    \item[] Justification: LLMs/VLMs are used for high-level planning, grasp selection, and keyframe generation, as described in Appendix~\ref{sec:interpolation} and the \emph{Following Text Instructions} appendix. These are components of the manipulation pipeline, rather than tools used only for manuscript preparation.

\end{enumerate}

\end{document}